%% file: main.tex
\pdfoutput=1
\documentclass[11pt]{article}
\usepackage[letterpaper,margin=1in]{geometry}
\usepackage[T1]{fontenc}
\usepackage{mathptmx}
\usepackage{microtype}
\usepackage{graphicx}
\usepackage{multirow}
\usepackage{amsmath,amssymb,amsfonts}
\usepackage{amsthm}
\usepackage{booktabs}
\usepackage{array}
\usepackage[authoryear,round]{natbib}
\usepackage[font=small,labelfont=bf]{caption}
\usepackage{authblk}
\usepackage{xcolor}
\usepackage[colorlinks=true,linkcolor=blue,citecolor=blue,urlcolor=blue]{hyperref}
\newcommand{\botrule}{\bottomrule}
\makeatletter
\g@addto@macro\@floatboxreset\centering
\makeatother

\theoremstyle{plain}%
\newtheorem{theorem}{Theorem}%
\newtheorem{proposition}[theorem]{Proposition}%
\newtheorem{hypothesis}{Hypothesis}%
\newcommand{\wg}{\mathrm{wg}}
\newcommand{\cov}{\mathrm{cov}}
\newcommand{\covgap}{\Delta_{\cov}}
\newcommand{\auroc}{\mathrm{AUROC}}
\newcommand{\Wone}{W_1}
\newcommand{\rhocal}{\rho_{\mathrm{cal}}}
\newcommand{\rhotest}{\rho_{\mathrm{test}}}

\begin{document}
\title{Calibration Beats Robust Training for Worst-Group Coverage}
\author[1]{One Octadion\thanks{Corresponding author: \texttt{octadion@student.ub.ac.id}}}
\author[1]{Novanto Yudistira}
\author[2]{Yuta Nakashima}
\affil[1]{Universitas Brawijaya, Malang, Indonesia}
\affil[2]{SANKEN, The University of Osaka, Osaka, Japan}
\date{}
\maketitle

\begin{abstract}
Group-robust training is evaluated almost entirely by worst-group accuracy, and it is
natural to assume that a model made accurate on its worst group will also give that
group reliable prediction sets: sets that contain the true label at the stated rate.
Across two spurious-correlation benchmarks and four backbones, we find instead that
worst-group coverage depends far more on how the scores are turned into sets than on how
the model was trained.
With one threshold shared by all groups, an ordinary model covers $89\%$ of all test
images but only $51\%$ of its worst group, and per-group thresholds close most of that gap
under every training method we tried, without changing the model or its scores. Fine-tuning the representation
end-to-end does not change this. Training tends to reduce the worst group's set size, but
only once calibration is group-conditional. If the worst group has to be reliable, the
calibration rule is the place to start.
\end{abstract}


\section{Introduction}
\label{sec:intro}

A waterbird on land is still a waterbird, yet a classifier trained on images in which
waterbirds nearly always appear over water will often misclassify it. Conformal prediction
\citep{vovk2005,angelopoulos2020,romano2020} is meant to make such a model's uncertainty explicit by
returning a set of labels, calibrated to include the true label with a stated probability. In one of
the settings we study, the sets contain the right species for $89\%$ of all test images, close to
the $90\%$ target, but for only $51\%$ of waterbirds on land. This does not violate the guarantee,
which is marginal and says nothing about any particular group \citep{cresswell2024}.

Two lines of work address this group-level failure, usually separately. Group-robust training
targets the worst group's accuracy, and group-conditional calibration targets coverage within each
group. They are linked by the rarely tested assumption that a representation made accurate on its
worst group will also make that group conformally reliable. If so, robust training should also
improve the uncertainty estimates built on it.

In our experiments, the assumption does not hold. When we hold the model fixed and change only the
calibration rule, worst-group coverage rises by up to $0.345$. Across our eight
backbone--dataset settings, we find that the gain from per-group thresholds is associated with the
Kolmogorov--Smirnov distance between the score distributions of the worst group and the remaining
groups.

Our contributions are:
\begin{enumerate}
\item \textbf{The calibration rule moves worst-group coverage far more than the training
method does (Section~\ref{sec:c1}).} With the adaptive prediction sets (APS) score
\citep{romano2020}, worst-group coverage in the eight settings varies across training methods by
up to $0.358$ under one shared threshold and by at most $0.024$ under per-group thresholds.
Fine-tuning the backbone end-to-end does not change this
(Section~\ref{sec:repr}).

\item \textbf{The size of that effect is bounded by the distance between the groups' score
distributions (Section~\ref{sec:c1}).} The shortfall under a shared threshold satisfies the
Kolmogorov--Smirnov bound of Eq.~\eqref{eq:ksbound} in every setting, tightly on Waterbirds and
loosely on CelebA. This is an empirical check of Proposition~\ref{prop:dissociation}.

\item \textbf{Training tends to reduce set size, but only once calibration is group-conditional
(Section~\ref{sec:c2}).} With coverage matched across methods, the best robust method gives the
worst group smaller sets than ERM in five of the six settings where ERM is kept. The per-setting
evidence is too weak to rank methods, and we report this as a tendency.

\item \textbf{Divergence can be compared at matched accuracy in only one of eight settings, and
there robust training lowers it (Section~\ref{sec:h1}).} Robust training moves base accuracy away
from ERM's. Without matching, a lower divergence cannot be separated from this accuracy effect, so
we match accuracy per training seed.
\end{enumerate}

\begin{figure}[t]
\centering
\includegraphics[width=\textwidth]{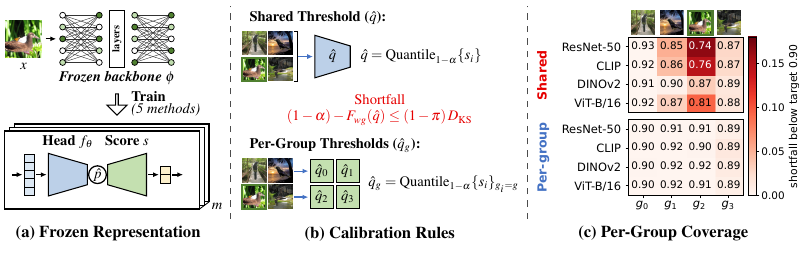}
\caption{\textbf{(a)} A head $f_\theta$ is trained on the features of a frozen backbone $\phi$ with each of the five methods, and its predicted probabilities $\hat p$ give the score $s$. \textbf{(b)} The two calibration rules: one shared threshold $\hat q$ for all groups, or a separate threshold $\hat q_g$ for each group. The shortfall under the shared threshold is bounded as in Eq.~\eqref{eq:ksbound}. \textbf{(c)} Coverage of each Waterbirds group under each rule, one row per backbone, averaged over the kept training methods (APS, $\rhocal=\rhotest=0.95$; Table~\ref{tab:pergroupcov}). Shading gives the shortfall below the target $0.90$. The minority groups $g_1$ and $g_2$ each hold $2.5\%$ of the calibration set.}
\label{fig:overview}
\end{figure}

Figure~\ref{fig:overview} shows the setup, the two calibration rules, and the coverage of each
Waterbirds group under each rule.

\section{Related Work}
\label{sec:related}

\paragraph{Group-robust training.} Worst-group accuracy is improved by
distributionally robust optimization \citep{sagawa2020}, last-layer retraining on
group-balanced data (DFR) \citep{kirichenko2023}, automatic feature reweighting
(AFR) \citep{qiu2023}, and pseudo-group methods such as JTT \citep{liu2021jtt} and
CNC \citep{zhang2022cnc}. The standard subpopulation-shift benchmark
\citep{yang2023change} compares these on accuracy-family metrics and calibration
error but reports no set-valued or coverage quantities. We take our last-layer methods from
this literature and measure what they do to conformal coverage and set size.
\citet{jones2021} show that selective classification can magnify accuracy disparities between
groups, and that on distributionally robust models it improves every group. We ask whether robust
training has the same effect on conformal coverage.

\paragraph{Group-conditional and fair conformal prediction.} Because exact feature-conditional
coverage is impossible distribution-free in finite samples
\citep{lei2014distribution,barber2021limits}, group-conditional coverage is the practical target. Mondrian conformal prediction
\citep{vovk2003mondrian,vovk2005} computes group-conditional thresholds, \citet{gibbs2023} and
\citet{jung2023} give conditional guarantees beyond disjoint groups, and
\citet{deng2025kandinsky} extend group-conditional coverage to overlapping and fractional
memberships. \citet{romano2020equalized} introduced equalized coverage as a fairness standard for
prediction sets. \citet{cresswell2024} show that equalizing coverage can increase disparate impact
and recommend equalizing set sizes, and \citet{afcp2024} adaptively equalize coverage while
controlling efficiency. \citet{burden2026} prove that equalized coverage and equalized set size
cannot in general hold simultaneously and recast the calibration rule as a choice of \emph{where}
cross-group heterogeneity manifests. These papers take the trained model as given. We vary the
\emph{training} intervention, and with it the representation and the score distribution, and ask
whether training reduces that cross-group heterogeneity in the scores themselves.

\paragraph{Conformal prediction under shift.}
\citet{tibshirani2019} establish validity under covariate shift through likelihood reweighting.
Multi-distribution robust conformal prediction \citep{mdcp2026} targets uniform worst-case coverage
as a group-DRO analogue \emph{inside the calibration procedure}, holding the model fixed, a
design consistent with our main finding. No prior work crosses the training interventions with calibration
rules under a correlation-strength sweep, nor controls for the accuracy confound we isolate.

\paragraph{Group recovery.} Biased ERM representations encode the spurious attribute strongly,
the principle behind JTT and CNC. Conformal prediction under noisy class or group labels
\citep{einbinder2022,sesia2023noisy} corrects coverage but assumes the group is given. We instead
keep the group \emph{recoverable} from the features used for calibration, where it remains
available for a later audit, and we test per-group thresholds with groups predicted from those
features (Section~\ref{sec:predgroup}).

\section{Problem Setup and Notation}
\label{sec:setup}

We observe triples $(x,y,a)$ with image $x$, label $y\in\mathcal{Y}$, and binary
spurious attribute $a\in\{0,1\}$ (the background, or \emph{place}, for Waterbirds and
\emph{Male} for CelebA). Groups are $g=(y,a)\in\mathcal{G}$,
$|\mathcal{G}|=2|\mathcal{Y}|$. A frozen backbone $\phi$ maps $x$ to features
$\phi(x)\in\mathbb{R}^d$, and a last-layer head $f_\theta$ trained with method $m$
maps $\phi(x)$ to class scores. The \emph{correlation strength} $\rho$ is the
probability that $a=y$ in the evaluation data. We calibrate at $\rhocal=0.95$ and evaluate over
$\rhotest\in\{0.95,0.9,0.8,0.7,0.6,0.5\}$, holding the calibration distribution
fixed.

\paragraph{Conformal sets and calibration rules.} Throughout, ``calibration''
means the conformal step that turns scores into a threshold, not the probability
calibration measured by expected calibration error \citep{guo2017calibration}. Given a nonconformity score
$s(x,y)$ and a calibration set $\mathcal{D}_{\mathrm{cal}}$, split conformal
prediction computes
$\hat q=\mathrm{Quantile}_{1-\alpha}\big(\{s(x_i,y_i)\}_{i\in\mathcal{D}_{\mathrm{cal}}}\big)$
and returns $C(x)=\{y\in\mathcal{Y}: s(x,y)\le\hat q\}$. We use three scores. THR is
$s_{\mathrm{THR}}(x,y)=1-\hat p(y\mid x)$, APS \citep{romano2020} is the cumulative
softmax mass up to $y$, and RAPS \citep{angelopoulos2020} is APS with a tail
regularizer. We compare three \emph{calibration rules}. \textbf{One shared threshold}, also called
marginal calibration or marginal split conformal, is the single global $\hat q$ above. \textbf{Per-group
thresholds}, also called Mondrian or group-conditional calibration, use a separate $\hat q_g$ for
each group. They admit a candidate label $c$ when $s(x,c)\le\hat q_{g(c)}$, where $g(c)$ is the
group that label would place the point in. The set can be computed without the test label. A third
rule, \textbf{shift-robust} calibration, inflates the threshold by a calibration-to-test shift
measured on test scores at their true labels. It cannot be deployed, and we report it only as an
upper bound. We target $1-\alpha=0.90$ throughout and give the full configuration, including the
shift-robust procedure, in Appendix~\ref{app:hyper}.

\paragraph{Reliability quantities.} For group $g$, $\cov_g=\Pr[y\in C(x)\mid g]$
and $|C|_g$ is its mean set size. We report worst-group coverage $\min_g\cov_g$
and the \emph{coverage gap}, i.e.\ the worst group's shortfall from the target,
\begin{equation}
\covgap=(1-\alpha)-\min_g\cov_g.
\label{eq:covgap}
\end{equation}
The gap is positive exactly when the worst group is under-covered. Table~\ref{tab:h2} reports $\covgap$
directly, and the other coverage tables report $\min_g\cov_g$. We also report mean set size and
the unweighted mean over groups $|\mathcal{G}|^{-1}\sum_g\cov_g$, the level that per-group
thresholds target. We need the mean because a minimum over a few groups is biased downward
(Appendix~\ref{app:subtarget}).

\paragraph{Cross-group score divergence.} Following the
conserved-quantity view of \citet{burden2026}, we summarize cross-group score
heterogeneity by a divergence between the worst group's score
distribution $P_{\wg}$ and that of the remaining groups $P_{\neg\wg}$. The bound in
Proposition~\ref{prop:dissociation} uses the Kolmogorov--Smirnov divergence
\begin{equation}
D_{\mathrm{KS}} = \sup_t\big|F_{\wg}(t)-F_{\neg\wg}(t)\big|,
\label{eq:div}
\end{equation}
computed for each of the three scores on a held-out split. Section~\ref{sec:h1} adds the
Wasserstein-1 divergence.

\paragraph{A dissociation result.}

\begin{proposition}[Calibration--representation dissociation]
\label{prop:dissociation}
Fix a score $s$ and target $1-\alpha$, and assume within-group exchangeability of
calibration and test scores.
\emph{(i)} Group-conditional (Mondrian) split conformal, with per-group threshold
$\hat q_g$ the $\lceil(1-\alpha)(n_g+1)\rceil$-th smallest calibration score in
group $g$, satisfies $\Pr\!\big(Y\in C(X)\mid G=g\big)\ge 1-\alpha$ for every group
$g$, with no dependence on $s$ beyond exchangeability.
\emph{(ii)} Let the scores be continuous, let $\pi=\Pr(G=\wg)$ under the calibration
distribution, and let $F_g$ be the group-$g$ nonconformity-score CDF. Marginal split
conformal uses the single pooled threshold $\hat q$ solving
$\pi F_{\wg}(\hat q)+(1-\pi)F_{\neg\wg}(\hat q)=1-\alpha$, and has per-group coverage
$\Pr\!\big(Y\in C(X)\mid G=g\big)=F_g(\hat q)$. The worst-group shortfall then obeys
the exact identity
\begin{equation}
(1-\alpha)-F_{\wg}(\hat q)\;=\;(1-\pi)\,\big[F_{\neg\wg}(\hat q)-F_{\wg}(\hat q)\big],
\label{eq:shortfall}
\end{equation}
and consequently
\begin{equation}
(1-\alpha)-F_{\wg}(\hat q)\;\le\;(1-\pi)\,D_{\mathrm{KS}}.
\label{eq:ksbound}
\end{equation}
Hence the representation (through $s$, and through any training intervention that
changes $s$) controls worst-group coverage \emph{only} under marginal calibration;
under Mondrian it does not. \end{proposition}

\noindent The proof is in Appendix~\ref{app:proofs}. \citet{jones2021} decompose the margin
distribution the same way to ask when abstention helps the worst group, with a confidence cut in
place of a conformal quantile and accuracy in place of coverage. Both the accuracy disparities
that abstention can magnify and the shortfall in part (ii) arise from a single threshold applied to
two different score distributions.

\paragraph{Runs and evaluations.} We call one (backbone, dataset, training method, seed) tuple a
\emph{run}, and one conformal evaluation of a run, at a given score, calibration split and
$\rhotest$, an \emph{evaluation}.

\noindent\textbf{Remark (limits of the divergences).} $D_{\mathrm{KS}}$ bounds the shortfall because
it is a supremum over $t$, but the Wasserstein-1 divergence admits no such bound, and the shortfall
need not grow with it. Either divergence orders coverage across models only when the same group is
worst in each. We therefore apply the identity and the bound in each evaluation, to whichever group
is worst there (Appendix~\ref{app:proofs}).

Part (i) holds by construction. We test part (ii) and ask how large the gap of
\eqref{eq:shortfall} is in practice and whether training shrinks it beyond what its accuracy gain
explains (Sections~\ref{sec:c1} and~\ref{sec:h1}). At a given CDF gap, the factor $(1-\pi)$ makes
the shortfall largest when the worst group is small. This fits Waterbirds, where the worst group
under a shared threshold holds $2.5\%$ of the calibration set. On CelebA, resampling to
$\rhocal=0.95$ leaves that group $47.5\%$, and the shortfall comes instead
from the cross-group CDF gap in \eqref{eq:shortfall}.

Part (i) holds in expectation over calibration draws, and the worst-group coverage we report under
per-group thresholds falls below $1-\alpha$ for finite-sample reasons (Section~\ref{sec:c1} and
Appendix~\ref{app:subtarget}).

\section{Experimental Setup}
\label{sec:exp}

\paragraph{Datasets and backbones.} We use Waterbirds \citep{sagawa2020} and CelebA
\citep{liu2015celeba}, where the task is blond-hair classification and \emph{Male} is the spurious
attribute. Both tasks are binary ($|\mathcal{Y}|=2$) with four groups $g=(y,a)$.
Table~\ref{tab:data} gives the split sizes, and we release per-group counts with our code. We
extract features with four frozen backbones that span two architecture families and three
pretraining regimes. They are an ERM-trained ResNet-50 \citep{he2016resnet} ($d{=}2048$, the
canonical setup), CLIP ViT-B/32 \citep{radford2021clip} ($d{=}512$, image--text contrastive),
DINOv2 ViT-B/14 \citep{oquab2024dinov2} ($d{=}768$, self-supervised and never label-trained), and a
supervised ImageNet ViT-B/16 \citep{dosovitskiy2021vit} ($d{=}768$). All methods use the same
extracted features. In Section~\ref{sec:repr}, we also fine-tune the ResNet-50 end-to-end under
three objectives to change the representation itself.

\paragraph{Training methods.} On frozen features we fit the head with five methods: ERM, DFR, AFR,
balanced subsampling \citep{idrissi2022balancing}, and last-layer GroupDRO (GroupDRO-LL).

\paragraph{Correlation strength.} For each target $\rho$, we resample the evaluation images into a
class-balanced set in which $a=y$ with probability $\rho$, and we record the empirical $\rho$ of
each draw. Before resampling, $a$ and $y$ are close to independent in the evaluation sets
($\rho\approx0.51$ on Waterbirds, $0.47$ on CelebA). On Waterbirds, $a=y$ matches the training
association (waterbirds over water). On CelebA, it reverses the raw association between blond hair
and being female.

\paragraph{Uncertainty.} We report means over training seeds, each with ten calibration splits. The
main grid evaluates every training method under one shared threshold, with five seeds. The
calibration ablation crosses every training method with three calibration rules, six correlation
strengths and three scores. It uses three seeds, because of that size and because two of its five
methods ignore the seed. The fine-tuning study uses five seeds on Waterbirds and three on CelebA,
whose training set is thirty-four times larger. Because the ten calibration splits inside a seed
share one trained model, intervals are two-stage cluster bootstraps ($B=2000$) that resample
training seeds and then the calibration splits within them. The correlation intervals of
Table~\ref{tab:c2} instead resample methods. Table~\ref{tab:variance} reports the seed and split
standard deviations separately.

\paragraph{Deterministic solvers.} ERM and AFR fit their heads with L-BFGS, whose solution does not
depend on the random seed (Appendix~\ref{app:hyper}). Extra seeds therefore add replicates for DFR,
balanced subsampling and GroupDRO-LL (across-seed SD up to $0.027$ in worst-group
accuracy) and none for ERM or AFR, whose SD is exactly zero in every setting.

\paragraph{Exclusions.} We exclude the $33$ of our $200$ runs whose worst-group accuracy falls below
a floor set for their training method (Appendix~\ref{app:excl} lists them and repeats the analysis
with them put back). \emph{Kept} runs are the others, and a method is kept in a setting unless all
of its runs there are excluded. ERM's worst-group accuracy spans $0.20$--$0.84$, and the robust
methods average $0.63$--$0.95$. We keep but flag twelve DFR runs on CelebA that fall between DFR's
floor and its soft reference, and we say so wherever an analysis leaves them out. We fixed the
floors and the six hypotheses of Appendix~\ref{app:hyps} before running the experiments they
concern, but did not lodge them in a public registry, so they are pre-specified rather than
preregistered.

\paragraph{Comparison with published values.} The gaps between our ResNet-50 methods and published
worst-group accuracies are large (Table~\ref{tab:lit}), and the comparison is not like-for-like in
two respects: we refit only the last layer on features from a deliberately simple training setup,
and we measure accuracy on our pooled evaluation set. Our main conclusions rest on comparisons
within a setting on the same evaluation data, and Appendix~\ref{app:hyper} discusses the gaps
method by method.

\begin{table}[t]
\centering
\caption{Worst-group accuracy of our ResNet-50 methods against published values (\citet{qiu2023}, Table~1). Ours average all five seeds, so on CelebA they include the three seeds excluded from the analysis, and AFR there is the degenerate method of Appendix~\ref{app:excl}. $^{\dagger}$The published numbers use group annotations; ours use them only to fit a last layer on frozen features.}
\label{tab:lit}
\small
\begin{tabular}{lcccc}
\toprule
 & \multicolumn{2}{c}{Waterbirds} & \multicolumn{2}{c}{CelebA} \\
Method & ours & published & ours & published \\
\midrule
ERM & 0.510 & 0.726 & 0.416 & 0.472 \\
DFR & 0.812 & 0.929$^{\dagger}$ & 0.851 & 0.883$^{\dagger}$ \\
AFR & 0.686 & 0.904 & 0.434 & 0.820 \\
GroupDRO (last-layer here) & 0.654 & 0.914$^{\dagger}$ & 0.757 & 0.889$^{\dagger}$ \\
Bal.\ subsample & 0.632 & --- & 0.753 & --- \\
\botrule
\end{tabular}
\end{table}

\section{Results}
\label{sec:results}

We ask what worst-group coverage depends on, and what training changes once the calibration rule is
fixed. Coverage depends mainly on the rule (Sections~\ref{sec:c1} and~\ref{sec:repr}), while
training tends to reduce set size and lowers divergence where accuracy can be matched
(Sections~\ref{sec:c2} and~\ref{sec:h1}). In Section~\ref{sec:h3}, we test deployment conditions.

\subsection{Calibration Rule versus Training Method}
\label{sec:c1}

\begin{figure}[t]
\centering
\includegraphics[width=\textwidth]{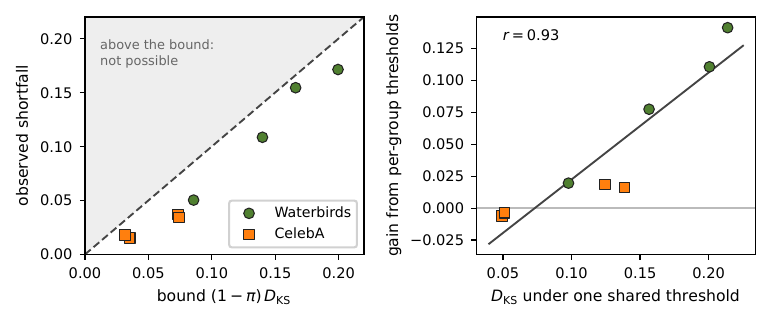}
\caption{\emph{Left:} the shortfall under one shared threshold against the bound $(1-\pi)D_{\mathrm{KS}}$ of Eq.~\eqref{eq:ksbound}. \emph{Right:} $D_{\mathrm{KS}}$, measured under the shared threshold, against the gain from per-group thresholds. Each point is one backbone--dataset setting, pooled over the kept runs except the ablation's seven flagged DFR runs on CelebA (APS, $\rhocal=\rhotest=0.95$).}
\label{fig:mech}
\end{figure}

\begin{figure}[t]
\centering
\includegraphics[width=\textwidth]{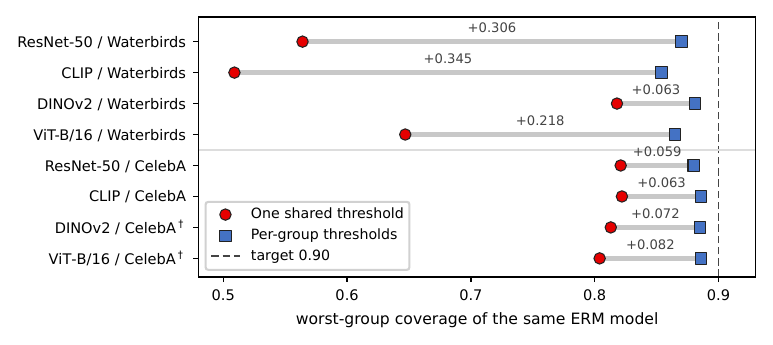}
\caption{Worst-group coverage of the same ERM model under one shared threshold (red) and under per-group thresholds (blue), one row per backbone--dataset setting, with the lift annotated. $^{\dagger}$ERM is below its accuracy floor here; values come from Appendix~\ref{app:excl}.}
\label{fig:c1}
\end{figure}

We first ask whether the calibration rule or the training method moves worst-group coverage more
when both act on the same frozen features. The rule moves it far more. Table~\ref{tab:c1} and Figures~\ref{fig:mech} and~\ref{fig:c1} come from the three-seed calibration
ablation. Across training methods, worst-group coverage varies by up to $0.358$ under one shared
threshold and by at most $0.024$ under per-group thresholds (Table~\ref{tab:c1}). We call this range
across a setting's kept training methods its spread, and this pattern the dissociation
(cf.~Proposition~\ref{prop:dissociation}). Switching only the calibration rule raises ERM's
worst-group coverage on all four backbones (Figure~\ref{fig:c1}), and the shortfall under one shared
threshold stays within the Kolmogorov--Smirnov bound of Eq.~\eqref{eq:ksbound}
(Figure~\ref{fig:mech}).

We call the per-group minus the shared coverage of the same ERM model, with nothing retrained, the
lift. Every interval for the lift excludes zero (Table~\ref{tab:c1}), and the lift is largest where
the shared threshold failed worst. On CLIP/Waterbirds, it takes ERM to $0.854$. DFR reaches $0.850$
there with one shared threshold, in the five-seed main grid (Table~\ref{tab:h2}).

The gain from per-group thresholds, pooled over training methods, can be anticipated from the score
distributions before the rule is switched. The bound $(1-\pi)D_{\mathrm{KS}}$ of
Eq.~\eqref{eq:ksbound} holds in all eight settings (Figure~\ref{fig:mech}, left), where we apply it
evaluation by evaluation to whichever group the shared threshold leaves worst. The bound is not
uniformly tight, and the shortfall reaches $0.59$--$0.93$ of the bound on Waterbirds but only
$0.43$--$0.58$ on CelebA. The bound is informative where the effect is large.

$D_{\mathrm{KS}}$, measured under the shared threshold, is also associated with the gain
(Figure~\ref{fig:mech}, right). Where the score distributions of the worst group and the remaining
groups nearly coincide, as on CelebA with DINOv2 and ViT-B/16, there is little to recover. The
eight points form two clusters, one per dataset, and the correlation printed in the figure has no
confidence interval. We treat the association as descriptive.

\paragraph{Further checks.} In an equivalence check, the upper end of the one-sided bootstrap
interval for the per-group spread falls below a margin of $0.05$ in all eight settings. That margin
was not pre-specified, and Appendix~\ref{app:supp} explains how we chose it and repeats the check
with a tighter one. With three seeds, the intervals in Table~\ref{tab:c1} indicate direction and
rough width, but their $95\%$ level is approximate. RAPS and THR reproduce the dissociation
(Appendix~\ref{app:supp}).

In the five-seed main grid, choosing the method with the highest worst-group accuracy rarely costs
coverage under one shared threshold. In seven of eight settings, that method's coverage gap
$\covgap$ is not significantly larger than the smallest (Table~\ref{tab:h2}). The two methods differ
in five settings, but a paired-bootstrap interval for the difference excludes zero only on
ResNet-50/CelebA.

Under per-group thresholds, worst-group coverage falls a little below the nominal $0.90$. Under APS
and averaged over the kept training methods, it is $0.856$--$0.884$ across the eight settings, and
individual methods range over $0.83$--$0.89$ across all three scores. The mean over
groups, which per-group thresholds target, is $0.9024$. The minimum over four noisy per-group
coverages is $0.018$--$0.045$ below that mean, about where an exactly valid procedure would put it.
Appendix~\ref{app:subtarget} gives the full argument and the comparison with that procedure, and
Appendix~\ref{app:hyps} tests this level against the pre-specified criterion.

\begin{table}[t]
\centering
\caption{Worst-group coverage under the two calibration rules (APS, $\rhocal=\rhotest=0.95$), with two-stage cluster-bootstrap intervals over training seeds and the calibration splits within them. ``Lift'' is the per-group minus shared coverage of the same ERM model; ``spread'' is the range across the kept training methods. Both are computed before rounding, so they can differ by $0.001$ from the printed columns. $^{\dagger}$ERM is below its accuracy floor here: its values come from the sensitivity run of Appendix~\ref{app:excl}, are left out of the spread, and are quoted without an interval.}
\label{tab:c1}
\footnotesize
\setlength{\tabcolsep}{3.5pt}
\begin{tabular}{lcclcl}
\toprule
 & \multicolumn{3}{c}{the same ERM model} & \multicolumn{2}{c}{spread across methods} \\
\cmidrule(lr){2-4}\cmidrule(lr){5-6}
Backbone & shared & per-group & lift [95\% CI] & shared & per-group [95\% CI] \\
\midrule
\multicolumn{6}{l}{\emph{Waterbirds}} \\
ResNet-50 & 0.564 & 0.870 & $+0.306$\,[$+0.292,+0.320$] & 0.290 & \textbf{0.008}\,[0.005,0.029] \\
CLIP & 0.509 & 0.854 & $+0.345$\,[$+0.322,+0.366$] & 0.358 & \textbf{0.024}\,[0.012,0.049] \\
DINOv2 & 0.818 & 0.881 & $+0.063$\,[$+0.052,+0.075$] & 0.049 & \textbf{0.024}\,[0.016,0.044] \\
ViT-B/16 & 0.647 & 0.865 & $+0.218$\,[$+0.201,+0.234$] & 0.217 & \textbf{0.007}\,[0.005,0.021] \\
\midrule
\multicolumn{6}{l}{\emph{CelebA}} \\
ResNet-50 & 0.821 & 0.880 & $+0.059$\,[$+0.046,+0.072$] & 0.071 & \textbf{0.003}\,[0.002,0.016] \\
CLIP & 0.822 & 0.886 & $+0.063$\,[$+0.054,+0.074$] & 0.059 & \textbf{0.003}\,[0.002,0.011] \\
DINOv2$^{\dagger}$ & 0.813 & 0.885 & $+0.072$ & 0.013 & \textbf{0.003}\,[0.001,0.013] \\
ViT-B/16$^{\dagger}$ & 0.804 & 0.886 & $+0.082$ & 0.003 & \textbf{0.009}\,[0.003,0.017] \\
\botrule
\end{tabular}
\end{table}

\begin{table}[t]
\centering
\caption{The accuracy-best and shortfall-best method per setting, from the five-seed main grid rather than the three-seed calibration ablation of Appendix~\ref{app:supp} (APS, $\rhocal=0.95$). $\covgap$ is measured under one shared threshold and given with its cluster-bootstrap CI. An inversion is \textbf{real} when a paired bootstrap of the difference excludes zero; parenthesized values are point estimates whose interval includes zero.}
\label{tab:h2}
\footnotesize
\setlength{\tabcolsep}{3.5pt}
\begin{tabular}{llcclcl}
\toprule
Backbone & acc-best & wg acc & $\covgap$ [CI] & shortfall-best & $\covgap$ & inversion \\
\midrule
\multicolumn{7}{l}{\emph{Waterbirds}} \\
ResNet-50 & DFR & 0.812 & 0.048\,[0.037,0.060] & DFR & 0.048 & no \\
CLIP & DFR & 0.838 & 0.050\,[0.040,0.062] & Bal.\ sub. & 0.039 & no ($+0.011$) \\
DINOv2 & DFR & 0.949 & 0.036\,[0.030,0.043] & Bal.\ sub. & 0.033 & no ($+0.003$) \\
ViT-B/16 & DFR & 0.877 & 0.035\,[0.028,0.044] & DFR & 0.035 & no \\
\midrule
\multicolumn{7}{l}{\emph{CelebA}} \\
ResNet-50 & DFR & 0.851 & 0.022\,[0.019,0.025] & GroupDRO-LL & 0.008 & \textbf{real} $+0.014$ \\
CLIP & GroupDRO-LL & 0.905 & 0.018\,[0.016,0.020] & GroupDRO-LL & 0.018 & no \\
DINOv2 & GroupDRO-LL & 0.846 & 0.015\,[0.012,0.018] & Bal.\ sub. & 0.014 & no ($+0.001$) \\
ViT-B/16 & GroupDRO-LL & 0.874 & 0.019\,[0.017,0.021] & Bal.\ sub. & 0.018 & no ($+0.001$) \\
\botrule
\end{tabular}
\end{table}

\subsection{Retraining the Representation}
\label{sec:repr}

The dissociation also holds when robust training changes the whole network, at least for the
ResNet-50. In the fine-tuning study, the spread across the three fine-tuned representations
on Waterbirds is $0.249$ under one shared threshold but $0.004$ under per-group thresholds
(Figure~\ref{fig:repr}; Table~\ref{tab:reprfull} in Appendix~\ref{app:supp}). The dissociation is
not an artifact of freezing the backbone.

We fine-tune the ResNet-50 end-to-end under three objectives (plain ERM, GroupDRO and group
reweighting), with five fine-tuning seeds on Waterbirds and three on CelebA. We hold the head fixed
at plain ERM, and only the representation moves. How much of a pretrained
network to unfreeze changes what the resulting representation encodes, and full
fine-tuning can distort pretrained features relative to linear probing
\citep{kumar2022finetuning,elmehdi2024unfreezing}. We unfreeze the whole network to let each
objective reshape the representation as much as it can. This makes the fine-tuning study the most
demanding test of the dissociation in our setup.

\paragraph{Manipulation check.} A representation-level experiment is informative only
if the intervention moved the representation, and we verify this first.
GroupDRO fine-tuning raises worst-group accuracy from $0.516$ to $0.738$ on
Waterbirds ($+0.222$) and from $0.396$ to $0.470$ on CelebA ($+0.074$), the largest
gain any of our representation-level objectives produces. DFR moves worst-group accuracy further
at the last layer (Table~\ref{tab:lit}). This experiment requires only a substantial change to the
representation, and GroupDRO fine-tuning produces one. Reweighting did not help ($0.466$ and
$0.364$, both a little below ERM). We keep reweighting in the comparison because it still changed
the representation, even though it did not raise worst-group accuracy.

\paragraph{Coverage under the two rules.} Under the shared threshold, worst-group coverage follows
worst-group accuracy (Figure~\ref{fig:repr}). Switching to per-group thresholds alone brings the
weakest representation up to the other two. CelebA repeats the pattern on a smaller scale, with a
spread of $0.019$ under one shared threshold and $0.001$ under per-group thresholds, because its
worst-group coverage under the shared threshold was already close to target. Even when fine-tuning
improves the representation, switching the calibration rule raises worst-group coverage more than
switching the representation does. The benefit of retraining appears in worst-group accuracy, not
in worst-group coverage.

\begin{figure}[t]
\centering
\includegraphics[width=\textwidth]{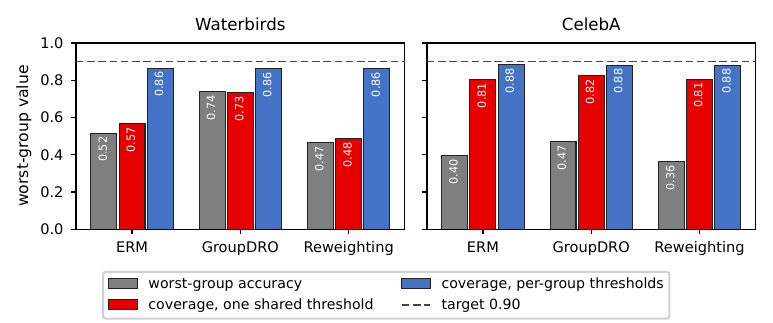}
\caption{Worst-group accuracy (gray) and worst-group coverage under one shared threshold (red) and under per-group thresholds (blue) for the three fine-tuned ResNet-50 representations, with the head held fixed at plain ERM (APS, $\rhocal=0.95$). Table~\ref{tab:reprfull} gives the values with their across-seed standard deviation.}
\label{fig:repr}
\end{figure}

\subsection{Set Size under Per-Group Thresholds}
\label{sec:c2}

If the calibration rule sets worst-group coverage, what does training still change? Once per-group
thresholds match worst-group coverage across methods, training tends to reduce the worst group's
set size. We call the robust method with the smallest worst-group sets the best robust method. In
the calibration ablation, it gives smaller sets than ERM in five of the six settings where ERM is
kept (Table~\ref{tab:c2}). Appendix~\ref{app:supp} names the methods with the smallest and largest
sets in each setting.

The per-setting evidence is too weak to rank methods. We compute each correlation between
worst-group accuracy and set size over three to five methods, and six of the eight bootstrap
intervals span essentially $[-1,+1]$. Two of these six, on CelebA, have coefficients near $-1$, but
with four methods such coefficients are not evidence of a deterministic relationship. The interval
for ViT-B/16 on CelebA is the only one that excludes zero, but it rests on three methods, too few
for its narrow width to indicate a precise estimate (Appendix~\ref{app:supp}). The interval for
ResNet-50/Waterbirds, where the correlation is positive, is narrower than the full range but still
contains zero. We therefore describe the effect of training on set size as a tendency. We state it for per-group thresholds only, because under one
shared threshold training still moves coverage.

\paragraph{Further checks.} Unlike coverage, worst-group set size is not shown to be equivalent
across training methods (Table~\ref{tab:c2}, spread column). The equivalence check of
Section~\ref{sec:c1}, applied with a wider margin of $0.10$, does not establish equivalence in six
of the eight settings. That margin was also chosen after seeing the spreads
(Appendix~\ref{app:supp}).

\begin{table}[t]
\centering
\caption{Worst-group set size under per-group thresholds, from the calibration ablation (APS, $\rhocal=\rhotest=0.95$). ``best robust'' is the robust method with the smallest worst-group sets. ``corr'' is the Pearson correlation between worst-group accuracy and set size over the setting's kept methods, with a bootstrap CI; \emph{italic} CIs span essentially $[-1,+1]$ or rest on three methods. $^{\dagger}$ERM excluded (Appendix~\ref{app:excl}). Sizes below $1$ indicate some empty sets.}
\label{tab:c2}
\footnotesize
\setlength{\tabcolsep}{4pt}
\begin{tabular}{lccccc}
\toprule
Backbone & ERM & best robust & spread & corr & CI \\
\midrule
\multicolumn{6}{l}{\emph{Waterbirds}} \\
ResNet-50 & 1.110 & 1.115 & 0.259 & $+0.35$ & $[-0.35,+1.00]$ \\
CLIP & 1.278 & 1.155 & 0.187 & $-0.54$ & \emph{$[-1.00,+1.00]$} \\
DINOv2 & 0.936 & 0.933 & 0.022 & $+0.15$ & \emph{$[-1.00,+1.00]$} \\
ViT-B/16 & 1.216 & 1.109 & 0.107 & $-0.80$ & \emph{$[-1.00,+1.00]$} \\
\midrule
\multicolumn{6}{l}{\emph{CelebA}} \\
ResNet-50 & 1.324 & 1.109 & 0.215 & $-0.97$ & \emph{$[-1.00,+1.00]$} \\
CLIP & 1.172 & 1.071 & 0.101 & $-1.00$ & \emph{$[-1.00,+1.00]$} \\
DINOv2 & $^{\dagger}$--- & 1.133 & 0.075 & $-0.51$ & \emph{$[-1.00,+1.00]$} \\
ViT-B/16 & $^{\dagger}$--- & 1.138 & 0.050 & $-0.85$ & \emph{$[-1.00,-0.85]$} \\
\botrule
\end{tabular}
\end{table}

\subsection{Divergence at Matched Accuracy}
\label{sec:h1}

Training may also change the score distributions themselves. Under a shared threshold,
Proposition~\ref{prop:dissociation} ties the shortfall to the gap between the worst group's scores
and those of the remaining groups. We therefore ask whether robust training narrows that gap by more
than its accuracy gain explains. It does, but only one setting of eight allows the comparison.

Here we measure the gap by the Wasserstein-1 divergence $D=\Wone(P_{\wg},P_{\neg\wg})$, the summary
used by \citet{burden2026}, computed like $D_{\mathrm{KS}}$ on a held-out split. Both $D$ and $\covgap$
in \eqref{eq:covgap} are entangled with base accuracy. A more accurate model produces sharper, more
group-aligned score distributions, and a lower $D$ under robust training partly reflects its
higher accuracy. A method's accuracy support is the range of its per-seed accuracies. For each
method $m$ and the ERM reference, we collect per-seed pairs
$\big(\mathrm{acc}^{(s)}_m,D^{(s)}_m\big)$ and evaluate the divergence at a common accuracy
$a^\star$ where the two supports \emph{overlap},
\begin{equation}
\Delta(a^\star)=D_{\mathrm{ERM}}(a^\star)-D_m(a^\star),
\label{eq:matched}
\end{equation}
with a $95\%$ two-stage cluster-bootstrap CI. A positive $\Delta(a^\star)$ means that $m$ lowers
the divergence at matched accuracy. Where the supports do not overlap, $\Delta(a^\star)$ is
\textbf{undefined} and reported as \texttt{unmatched}, and we report the raw $D$ separately as
uncontrolled.

In the five-seed main grid, robust training lowers the raw divergence $D$, but base accuracy moves
with it (Figure~\ref{fig:h1}; Table~\ref{tab:h1} in Appendix~\ref{app:hyper}). We can separate the two effects only on
ResNet-50/Waterbirds, where GroupDRO-LL lowers $D$ at matched accuracy, with
$\Delta_{\mathrm{APS}}=+0.084\,[+0.064,+0.101]$.

We match at model level, with one accuracy per training seed, because matching at split level
would compare ERM on unlucky evaluation draws against a robust method on lucky ones
(Appendix~\ref{app:hyper}). ERM's head is the same for every seed (Section~\ref{sec:exp}), and its
accuracy support is a single \emph{point}. Only GroupDRO-LL on ResNet-50/Waterbirds overlaps
it, and its per-seed accuracies bracket ERM's $0.9399$. Its reduction also holds under the other two
scores, with $\Delta_{\mathrm{RAPS}}=+0.084\,[+0.063,+0.101]$ and
$\Delta_{\mathrm{THR}}=+0.150\,[+0.129,+0.168]$, and all three two-stage cluster-bootstrap intervals
exclude zero.

We cannot match accuracy in the other seven settings. We exclude ERM in two. In the other
five, robust training moves accuracy away from ERM's. On the two CelebA settings among them, the
robust methods raise base accuracy above ERM's, with no common operating point
(Table~\ref{tab:h1}). On the three Waterbirds settings, no method's accuracy support contains ERM's
accuracy (Appendix~\ref{app:hyper}).

\begin{figure}[t]
\centering
\includegraphics[width=\textwidth]{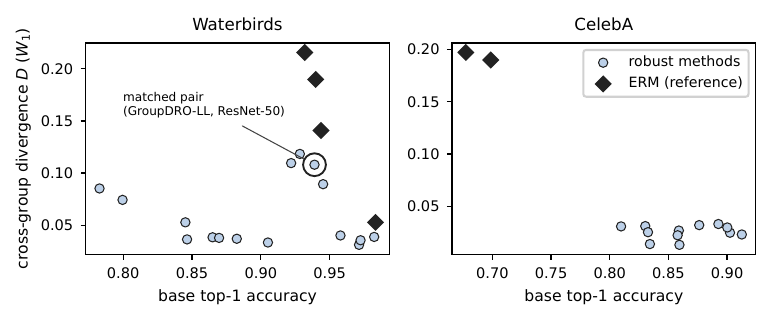}
\caption{Each point is one method's mean base accuracy against its cross-group divergence $D$ (APS, $\rhocal=0.95$); black diamonds mark each setting's ERM reference. Matching is within a setting, and the one pair in the grid with overlapping per-seed accuracy support is circled.}
\label{fig:h1}
\end{figure}

\subsection{Correlation Shift and Deployment}
\label{sec:h3}

\begin{figure}[t]
\centering
\includegraphics[width=\textwidth]{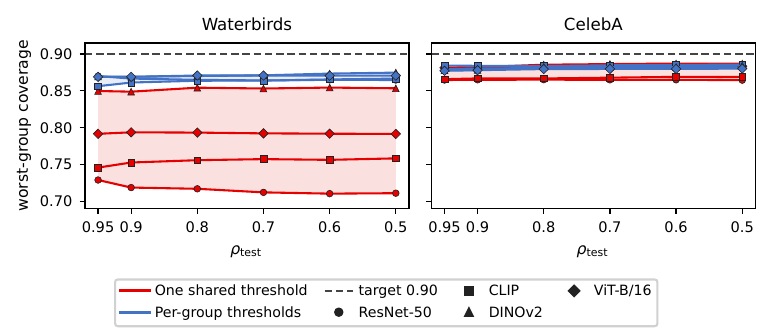}
\caption{Worst-group coverage against $\rhotest$, with calibration held at $\rhocal=0.95$ and coverage averaged over the kept training methods (APS). One curve per backbone, shaded to the envelope across the four; blue is per-group thresholds, red is one shared threshold.}
\label{fig:shift}
\end{figure}

Deploying per-group thresholds raises three questions that the comparisons above leave open:
whether coverage holds when the correlation strength $\rho$ shifts, whether the protected group can
still be audited, and whether the thresholds work without group labels. The coverage results hold
under all three conditions, the last for APS and RAPS only. In the calibration ablation,
worst-group coverage stays flat as $\rhotest$ moves away from $\rhocal=0.95$ (Figure~\ref{fig:shift}). Under APS with per-group thresholds, it varies by at most $0.024$ for every
method and setting.

With one shared threshold, worst-group coverage varies by at most $0.037$. The
sweep changes only the mixture of groups, and a change in the mixture leaves each group's coverage
under a fixed threshold unchanged. Because the curves are flat, the sub-target level of per-group
thresholds ($0.84$--$0.89$ for individual methods under APS; Section~\ref{sec:c1} and
Appendix~\ref{app:subtarget}) does not fall under shift. Appendix~\ref{app:supp} gives the spread across backbones and the
shift-robust rule.

\paragraph{Further checks.} The $\rho$ sweep changes the
\emph{group mixture} of the evaluation distribution while the class-conditional feature
distribution within each of the four groups is untouched. It is therefore a
group-prior shift, a special case of label-and-attribute shift, and \emph{not} a
general covariate shift. Our stability claim is scoped accordingly. We claim that per-group
thresholds are insensitive to the strength of the spurious correlation at deployment, and we do not
test within-group distribution change. On CelebA, where $\rho=0.95$ reverses the raw association
(Section~\ref{sec:exp}), lowering $\rhotest$ moves the evaluation distribution toward the raw data.

The advantage of per-group thresholds over one shared threshold
lies in the \emph{level} they attain, not in their variance. Under one shared threshold, the
split-to-split SD is $0.007$--$0.043$, comparable to that under per-group thresholds, and in five of
the eight settings per-group thresholds are the less stable rule (Table~\ref{tab:variance}).

\paragraph{Group auditing.}\label{sec:audit} We find that the deployed features do not hide the
protected group. To test this, we train a probe $h:\phi(x)\mapsto a$ in-domain to recover the
spurious attribute and measure its test $\auroc$. The probe is a logistic regression on
standardized features and never sees a test label. It recovers the attribute in all eight settings
(probe $\auroc$ in Table~\ref{tab:predgroup}), and the group can still be audited after worst-group
coverage is obtained. We also ask whether removing the attribute from the features changes
worst-group coverage. Projecting out the top-$k$ spurious-predictive directions
($k\in\{0,1,5,20\}$), for ResNet-50 and CLIP on both datasets, leaves the probe at
$\auroc\in[0.49,0.64]$ for $k=20$. Worst-group coverage moves by less than $0.005$ on CelebA and
by at most $0.022$ overall. The largest move is on CLIP/Waterbirds, from $0.854$ to $0.876$.
This exceeds the $0.02$ bar at which our test flags a tradeoff: on CLIP/Waterbirds, keeping the
attribute recoverable costs $0.022$ of worst-group coverage. Even with the attribute removed,
coverage there stays $0.024$ short of target. Removing the attribute barely changes worst-group coverage, and per-group thresholds already reach that coverage
with features from which the attribute is fully recoverable.

\paragraph{Predicted groups.}\label{sec:predgroup} Per-group thresholds also work, under APS and RAPS, when the groups must be
predicted because a deployment lacks group labels. In the predicted-group study, we fit a probe on
the training split's features to predict the attribute $\hat a$, and we never predict the label
$y$. Calibration points, whose labels are known, are binned by $\hat g=2y+\hat a$. A test point
admits candidate label $c$ when its score is at most $\hat q_{2c+\hat a}$, and the predicted
attribute is the only group information used at test. Because we always measure coverage against
the \emph{true} group, mis-assignments lower the reported coverage. Attribute labels are needed only
on the training split. Calibration still needs $y$, as in any split-conformal method, and the
requirement is weakened but not removed.

Under APS, worst-group coverage with predicted groups stays within $0.02$ of that with true groups
in $30$ of the $34$ (setting, method) pairs (Table~\ref{tab:predgroup}). RAPS gives the same count
and THR a lower one, and we restrict the claim to APS and RAPS (Appendix~\ref{app:supp}).
Mis-assignment costs are asymmetric by construction. Because the minority group has the largest
threshold, a minority point placed in a majority group is under-covered, while the reverse error
only enlarges its set. The largest differences come from ERM and then GroupDRO-LL. Probe accuracy
does not predict which pairs exceed $0.02$. ResNet-50/Waterbirds, the setting with the lowest probe
$\auroc$ ($0.945$), has no such pair, whereas CLIP/CelebA, at $\auroc=0.999$, has one
(Table~\ref{tab:predgroup}). A claim that predicted groups can replace true groups should therefore
report the probe's $\auroc$, but a high $\auroc$ alone does not support it.

\paragraph{Ethical scope.} The predicted-group version predicts a protected attribute for
individuals. CelebA's \texttt{Male} field is a binary annotation applied by third parties to
photographs, not self-identified gender, and the group it induces should be read only as ``the annotation that
co-varies with hair color in this dataset''. Inferring such an attribute at test time also differs
from conditioning on one a person has disclosed. It assigns a sensitive label to someone who has not
provided it, and for some of them it assigns the wrong one, even at the $\auroc$ we measure. We
recommend the predicted-group version only where the attribute is already lawfully held,
or where the inference stays inside an audit an operator runs on their own system and never reaches
a per-person decision. It is not appropriate where the inferred label would be stored, disclosed or
used to route individual outcomes. There, the true-group version should be used, with attributes
obtained by consent. Regimes governing special-category data may prohibit the inference outright,
whatever its accuracy.

\begin{table}[t]
\centering
\caption{Worst-group coverage with probe-predicted groups at both calibration and test, compared with ground-truth groups, across the four backbones (APS, $\rhocal{=}\rhotest{=}0.95$). ``diff.'' is $|\cov(a)-\cov(c)|$ between true groups (a) and predicted groups at both stages (c), over the setting's kept methods; ``deploy.'' counts methods within $0.02$; the last column names the method with the largest difference where any exceeds $0.02$. Table~\ref{tab:pgfull} gives all three conditions for APS and THR.}
\label{tab:predgroup}
\footnotesize
\setlength{\tabcolsep}{3.5pt}
\begin{tabular}{lccccl}
\toprule
 & & \multicolumn{2}{c}{diff.} & & \\
\cmidrule(lr){3-4}
Backbone & probe $\auroc$ & median & max & deploy. & worst method \\
\midrule
\multicolumn{6}{l}{\emph{Waterbirds}} \\
ResNet-50 & 0.945 & 0.007 & 0.012 & 5/5 & --- \\
CLIP & 0.976 & 0.008 & 0.051 & 4/5 & ERM \\
DINOv2 & 0.961 & 0.010 & 0.024 & 4/5 & ERM \\
ViT-B/16 & 0.964 & 0.006 & 0.022 & 4/5 & AFR \\
\midrule
\multicolumn{6}{l}{\emph{CelebA}} \\
ResNet-50 & 0.962 & 0.003 & 0.003 & 4/4 & --- \\
CLIP & 0.999 & 0.011 & 0.025 & 3/4 & GroupDRO-LL \\
DINOv2 & 0.999 & 0.012 & 0.013 & 3/3 & --- \\
ViT-B/16 & 0.993 & 0.001 & 0.003 & 3/3 & --- \\
\botrule
\end{tabular}
\end{table}

\section{Discussion}
\label{sec:discussion}

The calibration rule moves worst-group coverage far more than training does, and per-group
thresholds keep that coverage under correlation shift and, for APS and RAPS, with predicted groups.
We therefore recommend per-group thresholds wherever the worst group's coverage matters. The
remaining question for practice is what they cost.

The recommendation has one cost. \citet{cresswell2024} give
prediction sets to human decision-makers and find that equalizing coverage produces the
\emph{largest} outcome disparity of any rule they test, because it widens the gap in set sizes. Our
data reproduce this mechanism. Per-group thresholds raise the set-size disparity
$\max_g|C|_g-\min_g|C|_g$, the difference between the largest and the smallest mean set size over
groups, in six of the eight settings (Table~\ref{tab:pergroupcov}). On CLIP/Waterbirds
under ERM, the worst group's sets grow from $1.080$ to $1.278$ and the average from $0.984$ to
$1.214$.

Which rule to use depends on the application. Where coverage must be guaranteed, as in an audit, a
service-level agreement or a regulatory floor, we recommend per-group thresholds with their cost in
set size. That cost differs between the datasets. The average set size grows in all four Waterbirds
settings and shrinks in all four CelebA settings (Appendix~\ref{app:supp}). Where a person acts on
the set, the decision depends on its size, and equalizing coverage changes that size.

We draw two further recommendations for practice.
\begin{itemize}
\item Compare cross-group divergence \emph{at matched accuracy}, because an uncontrolled comparison
mixes the effect of training with the effect of accuracy. In the one setting where matching is
possible, the controlled estimate is at least as large as the uncontrolled one.
\item Under one shared threshold, a method chosen for worst-group accuracy is rarely a poor choice
for the coverage shortfall in our grid.
\end{itemize}

\section{Limitations}
\label{sec:limitations}

Our evidence has six limitations.
\begin{itemize}
\item We use two vision benchmarks with binary labels and binary spurious attributes, four frozen
backbones and one backbone fine-tuned end-to-end. With two classes a set holds at most two labels.
This compresses the efficiency axis, although set size still varies across training methods
(Table~\ref{tab:c2}). A third dataset and a multi-class task would
show whether the dissociation depends on $|\mathcal{Y}|=2$.
\item We move the representation for one architecture only. The fine-tuning evidence is narrower
than the frozen grid's.
\item We can compare divergence at matched accuracy in only one of the eight settings.
\item Coverage under per-group thresholds is close to a per-group target by construction. The
non-trivial findings are the \emph{magnitude} of the shared threshold's
training-dependent failure and of the lift in ERM's worst-group coverage from per-group thresholds.
\item The predicted-group version (Section~\ref{sec:predgroup}) degrades under THR, where nine of
thirty-four (setting, method) pairs miss, and it infers a protected attribute at test time.
\item Our confidence intervals do not model cross-dataset variance.
\end{itemize}

\section{Conclusion}
\label{sec:conclusion}

In our experiments, worst-group coverage depends far more on the calibration rule than on how the
model is trained. Per-group thresholds lift an ordinary model's worst-group coverage by up to
$0.345$, and under them the training method changes it by at most $0.024$ (APS). Robust training
remains useful. It tends to reduce the worst group's set size, but only under per-group
thresholds. We release the matched-accuracy protocol, with which others can check training-side conformal claims without the
accuracy confound wherever accuracies overlap.

\section*{Data and code availability}
Waterbirds and CelebA are public. We release the records behind every table---about
$116{,}000$ conformal evaluations in total, covering the main grid, the calibration
ablation, the predicted-group study and the fine-tuning study---together with the
log of which runs were excluded and why, and the scripts that draw each figure, at
\url{https://github.com/octadion/worst-group-conformal}, archived at
\url{https://doi.org/10.5281/zenodo.22762874}. The conformal evaluations re-run
in minutes from the stored scores, with no GPU retraining.

\section*{Declarations}
\textbf{Funding.} The authors received no specific funding for this work.
\textbf{Competing interests.} The authors have no competing interests relevant to
this article. \textbf{Ethics approval.} Not applicable for human-subjects review. This study uses only public benchmark datasets. Section~\ref{sec:predgroup} discusses the ethics of inferring a protected attribute at test time and scopes the recommendation accordingly. \textbf{Data and
code availability.} See ``Data and code availability'' above. \textbf{Author
contributions.} O.O.\ designed and ran the experiments and wrote the manuscript.
N.Y.\ and Y.N.\ supervised the work and revised the manuscript. All authors approved
the final version.

\appendix
\setcounter{figure}{0}
\setcounter{table}{0}
\setcounter{equation}{0}
\renewcommand{\thefigure}{\thesection\arabic{figure}}
\renewcommand{\thetable}{\thesection\arabic{table}}
\renewcommand{\theequation}{\thesection\arabic{equation}}
\renewcommand{\theHfigure}{appendix.\thefigure}
\renewcommand{\theHtable}{appendix.\thetable}
\renewcommand{\theHequation}{appendix.\theequation}
\section{Derivations}
\label{app:proofs}

\subsection*{Proposition~\ref{prop:dissociation}}

\begin{proof}
(i) is the standard finite-sample split-conformal guarantee applied within each
group under within-group exchangeability. For (ii), the group-$g$ coverage is
$F_g(\hat q)$ by definition of $C$. Writing the pooled CDF as the mixture
$F=\pi F_{\wg}+(1-\pi)F_{\neg\wg}$ and using $F(\hat q)=1-\alpha$,
\[
1-\alpha=\pi F_{\wg}(\hat q)+(1-\pi)F_{\neg\wg}(\hat q)
        =F_{\wg}(\hat q)+(1-\pi)\big[F_{\neg\wg}(\hat q)-F_{\wg}(\hat q)\big],
\]
which rearranges to \eqref{eq:shortfall}. Inequality \eqref{eq:ksbound} follows since
$F_{\neg\wg}(\hat q)-F_{\wg}(\hat q)\le\sup_t|F_{\wg}(t)-F_{\neg\wg}(t)|=D_{\mathrm{KS}}$.
\end{proof}

\subsection*{Worst-group identity}

Part (ii) holds for any designated group. Nothing that rests on it therefore requires the worst
group to be the same one across models. Neither divergence orders coverage across models without
assuming the same group is worst in each, and we do not assume it. The identity
\eqref{eq:shortfall} and the bound \eqref{eq:ksbound} are applied evaluation by evaluation to
whichever group is in fact worst. The assumption is a good approximation for the ERM baseline under
a shared threshold, where one group is the worst in $85$--$100\%$ of evaluations, and a poor one
under per-group thresholds, where no group holds the minimum in more than $41\%$ of evaluations and
the most concentrated setting is closer to the $25\%$ of a four-way tie than to the
shared-threshold figure (Table~\ref{tab:argmin}).

Under one shared threshold, the ERM baseline's worst group is not the same \emph{kind} of group in
the two datasets. On Waterbirds it is
waterbird-on-land, which holds $2.5\%$ of the calibration set. The small-$\pi$ reading of
\eqref{eq:shortfall} applies, and the group is a minority in the sense of the spurious-correlation
literature. On CelebA it is blond-and-male, one of the two label-attribute-aligned groups to which
resampling to $\rhocal=0.95$ gives $47.5\%$ each. This group is a minority in the \emph{class} but
not in the group prior. The shortfall there is therefore not a small-$\pi$ effect, and the mechanism
paragraph of Section~\ref{sec:setup} describes Waterbirds alone.

Under per-group thresholds, the most frequent worst group holds the minimum in $30$--$41\%$ of
evaluations. This is above the $25\%$ of a four-way tie, and therefore not a tie, but the minimum
is no longer pinned to one group. For this reason we report the level of worst-group coverage in
Section~\ref{sec:c1}, not which group attains it. The minimum of four noisy coverages also sits below
their mean, by the amount computed next.

\begin{table}[!ht]
\caption{Where the worst group falls. For the ERM baseline under one shared threshold: the group attaining the minimum, its share of the calibration set, and the percentage of evaluations in which it is worst. The last column is for the per-group evaluations of all kept methods: the percentage of evaluations in which the most frequent worst group is worst, where $25\%$ would be a four-way tie. Counts combine all scores and values of $\rhotest$. A dash marks a setting where ERM is below its accuracy floor (Appendix~\ref{app:excl}).}
\label{tab:argmin}
\footnotesize
\setlength{\tabcolsep}{5pt}
\begin{tabular}{@{}lcccc@{}}
\toprule
 & \multicolumn{3}{c}{ERM, one shared threshold} & per-group \\
\cmidrule(lr){2-4}\cmidrule(lr){5-5}
Setting & worst group & its share of cal.\ set & \% of evaluations & most frequent group, \% \\
\midrule
\multicolumn{5}{@{}l}{\emph{Waterbirds}} \\
ResNet-50 & $g_{2}$ & 0.025 & 100 & 37 \\
CLIP & $g_{2}$ & 0.025 & 99 & 41 \\
DINOv2 & $g_{2}$ & 0.025 & 85 & 33 \\
ViT-B/16 & $g_{2}$ & 0.025 & 94 & 30 \\
\multicolumn{5}{@{}l}{\emph{CelebA}} \\
ResNet-50 & $g_{3}$ & 0.475 & 100 & 37 \\
CLIP & $g_{3}$ & 0.475 & 100 & 31 \\
DINOv2 & --- & --- & --- & 37 \\
ViT-B/16 & --- & --- & --- & 35 \\
\botrule
\end{tabular}
\end{table}

\subsection*{Expected minimum over groups}

Appendix~\ref{app:subtarget} uses the approximation
$\mathbb{E}[\min_g\cov_g]\approx(1-\alpha)-c_{|\mathcal{G}|}\,\sigma$, in which $c_k$ is minus the
expected minimum of $k$ standard normals: $-\mathbb{E}[\min_{j\le k}Z_j]=k\int
z\,\phi(z)\Phi(z)^{k-1}\,dz$, giving $c_2=0.564$, $c_3=0.846$, $c_4=1.029$ and $c_5=1.163$.
With four groups the reported minimum therefore sits about one per-group standard deviation below
the mean over groups even when every group is exactly valid. Treating the per-group coverages as
independent and Gaussian is an approximation, because they share a test set and coverage is an
average of Bernoulli draws. The constant therefore only sets the scale, and the value we compare
against comes from the simulation of Table~\ref{tab:subtarget}.

\subsection*{A construction for $\Wone$}

The shortfall in \eqref{eq:shortfall} is the cross-group CDF gap \emph{at the threshold}, scaled by
$(1-\pi)$, and $D_{\mathrm{KS}}$ bounds it because $D_{\mathrm{KS}}$ is a supremum over $t$.
$\Wone$ admits no such bound: the gap can concentrate on a vanishing interval, sending $\Wone$ to
zero while the shortfall stays fixed. It therefore does not follow from \eqref{eq:shortfall} that
the shortfall grows with $\Wone$, and in general it does not. To construct such a case, fix the
shared threshold $\hat q$ and a target gap $\delta$, and let
$F_{\neg\wg}(t)-F_{\wg}(t)$ equal $\delta$ at $t=\hat q$ but fall to zero outside
$[\hat q-\varepsilon,\hat q+\varepsilon]$. Then
\begin{equation}
\begin{aligned}
\Wone(P_{\wg},P_{\neg\wg})
&= \int\big|F_{\wg}(t)-F_{\neg\wg}(t)\big|\,dt \\
&\le 2\varepsilon\delta \;\xrightarrow[\varepsilon\to0]{}\; 0 ,
\end{aligned}
\label{eq:w1zero}
\end{equation}
while the shortfall stays at $(1-\pi)\delta$ throughout. $\Wone$ can therefore be made arbitrarily
small at fixed shortfall, because it integrates the CDF gap over the line while the shortfall
depends on it at a single threshold. We therefore use $\Wone$ only as a descriptive summary, its
role in \citet{burden2026}. We report it in Table~\ref{tab:div} and in Section~\ref{sec:h1}, and we
use $D_{\mathrm{KS}}$ wherever we claim a bound or a prediction.

\section{Additional Results}
\label{app:supp}

This appendix gives the numbers behind the claims of Section~\ref{sec:results}. We compute its
tables from the released records, which give each (setting, method) pair separately.

\paragraph{The calibration ablation.}
Table~\ref{tab:scorespread} shows the dissociation under all three scores, THR with a slightly
larger per-group spread on average. The per-group spread is
the smaller of the two in $22$ of the $24$ (setting, score) pairs, and both exceptions are ViT-B/16
on CelebA, where the shared threshold already leaves almost nothing to recover.

Per-group thresholds lower worst-group coverage in seven of the thirty-four kept (setting, method)
pairs of Section~\ref{sec:c1}, each time by at most $0.014$ and where the shared threshold was
already near target. Under one shared threshold, coverage over all test images on CLIP/Waterbirds is
$0.892$ (SD $0.012$ over ten calibration splits), within one standard deviation of the target.

We chose the equivalence margin of $0.05$ in Section~\ref{sec:c1} after seeing the spreads, as the
largest coverage difference we would consider immaterial in deployment.
It is also wide relative to the spreads, and the check bounds the effect but does not resolve it.
If we tighten the margin to $0.04$, equivalence fails only on CLIP/Waterbirds, whose upper bound is
$0.044$, and DINOv2/Waterbirds clears it by less than $0.001$. For set size
(Section~\ref{sec:c2}), we chose the wider margin of $0.10$ in the same way. It favors equivalence,
and the check still fails to establish it in six of the eight settings.

Shift-robust calibration, the third rule, inflates its threshold with the observed
calibration-to-test distance and lifts most robust methods above target. Under APS, it raises
worst-group coverage by up to $0.347$ as $\rhotest$ falls. At
$\rhotest=0.95$, their worst-group coverage spans $0.84$--$0.97$. The four of their twenty-eight
(setting, method) pairs below $0.90$ are all on Waterbirds (GroupDRO-LL at $0.844$, $0.865$ and
$0.865$, balanced subsampling at $0.888$). ERM stays at $0.653$, $0.777$ and $0.798$ on three
Waterbirds settings, and the range over all kept (setting, method) pairs is therefore
$0.65$--$0.97$. Under APS and RAPS, the rule shows how much a shift-robust threshold could recover
when $\rhotest$ moves far from $\rhocal$. Like per-group thresholds, it makes worst-group coverage
much less dependent on training. These values are an upper bound, because the ball radius comes
from test scores (Section~\ref{sec:setup}).

Under THR, the shift-robust rule degenerates on CelebA. At $\rhotest=0.95$, twelve of CelebA's
fourteen kept (setting, method) pairs reach a mean worst-group set size above $1.9$, and seven sit at
$2.000$. With two classes, every set then holds both labels, and the coverage of $1.000$ is that of
the trivial predictor. We therefore scope the shift-robust recommendation to APS and RAPS and treat
its THR behavior on CelebA as a failure mode.

\paragraph{Divergences.} Table~\ref{tab:div} reports $D_{\mathrm{KS}}$ over all kept runs, together
with $\Wone$ for APS. Figure~\ref{fig:mech} leaves out the seven flagged DFR runs on CelebA, which
raises $D_{\mathrm{KS}}$ on CelebA/ResNet-50 from $0.130$ to $0.139$ and changes no other setting by
more than $0.005$. In the five-seed main grid under APS, the Wasserstein-1 divergence $D$ of
Section~\ref{sec:h1} falls from $0.05$--$0.22$ for ERM to $0.01$--$0.12$ for the robust methods
(Table~\ref{tab:h1}).

\paragraph{The correlation-strength sweep.} Averaged over the kept methods, as in
Figure~\ref{fig:shift}, no setting's worst-group coverage departs from its $\rhotest=0.95$ value by
more than $0.019$ under either rule. The largest departure is on ResNet-50/Waterbirds under the
shared threshold, where coverage is already far below target at $\rhocal$. Lowering $\rhotest$
also does not produce the ``relocate, not remove'' effect of \citet{burden2026}. From
$\rhotest=0.95$ to $0.5$, set-size disparity under APS grows in seven of the thirty-four kept
(setting, method) pairs. The increases are at most $0.011$, and the decreases at most $0.034$.
At $\rhotest=0.95$, the range across the four backbones on Waterbirds is $0.014$ under per-group
thresholds, against $0.121$ under one shared threshold. Part of the shared-threshold range comes
from our ResNet-50, which sets its low end and is trained with a simpler setup than published ones
(Section~\ref{sec:exp}).

\paragraph{The fine-tuning study.} Table~\ref{tab:reprfull} gives the fine-tuning study of
Section~\ref{sec:repr} in full, with the across-seed standard deviation and THR. The spreads quoted
there are computed before rounding, so they can differ by $0.001$ from the printed columns. On
Waterbirds, THR's worst-group coverage falls a further
$0.11$--$0.25$ below APS under the shared threshold and is unchanged under per-group thresholds.

\paragraph{Set size by method.} Under per-group thresholds, the most accurate method gives the
smallest worst-group sets in four of eight settings and comes within $0.010$ of the smallest in
three of the other four. A robust method does not always give smaller sets
than ERM. ERM gives the largest sets in three of its six settings, and in the other three a robust
method does: AFR on ResNet-50/Waterbirds and CLIP/Waterbirds ($1.369$ and $1.342$) and GroupDRO-LL
on DINOv2/Waterbirds ($0.954$). ResNet-50/Waterbirds is the only setting where ERM gives the
smallest sets, by $0.005$, and the only one where the most accurate method misses the smallest by
more than $0.010$, at $0.070$.

On DINOv2/Waterbirds, the mean worst-group set size is \emph{below} one for every training method
and every score ($0.890$--$0.954$), and a small fraction of prediction sets are empty. This is
expected for split conformal with a highly accurate model, since the calibration quantile can fall
below every class score. Worst-group coverage there is $0.857$--$0.882$ under per-group thresholds,
in line with the other settings. We do not clip sets to a minimum size of one, because clipping
would hide the empty sets.

\paragraph{Correlation intervals.} The correlation intervals of Table~\ref{tab:c2} resample the
methods of a setting. With three methods, most bootstrap resamples contain only two distinct points,
whose correlation is $\pm1$. The interval for ViT-B/16 on CelebA is narrow because of these
degenerate resamples, and its width does not indicate a precise estimate.

\paragraph{Set size under the two rules.} Switching from one shared threshold to per-group
thresholds changes set size differently on the two datasets (Section~\ref{sec:discussion}). On
Waterbirds, the average set size grows in all four settings, by up to $0.151$, and the worst group's
sets grow in three. On CelebA, where the worst group is one of the two large aligned groups, the
average shrinks in all four settings and the worst group's sets shrink in three.

\paragraph{Predicted groups.} Table~\ref{tab:pgfull} gives worst-group coverage under all three
conditions: (a) true groups, (b) predicted groups at test only and (c) predicted groups at
calibration and test. APS and RAPS behave alike, and THR is the least stable, because small changes
in the top-$1$ softmax mass matter more when a point is mis-assigned. Under APS, three of the four
(setting, method) pairs that differ by more than $0.02$ between (a) and (c)
(Table~\ref{tab:predgroup}) are on Waterbirds, whose evaluation set is nearly six times smaller
than CelebA's.

\paragraph{Variance decomposition.} Table~\ref{tab:variance} separates the standard deviation
across training seeds from that across calibration splits. This separation motivates the two-stage
cluster bootstrap we use throughout.

\paragraph{Coverage by group.} Table~\ref{tab:pergroupcov} breaks the same evaluations down by
group, from a re-run (APS, $\rhocal=\rhotest=0.95$) on the released splits. Its minimum over groups
matches the reported worst-group coverage in all but $305$ of the $2{,}400$ evaluations, all from
ERM, AFR or balanced subsampling (Appendix~\ref{app:hyper}).

Under per-group thresholds, every group's coverage lies between $0.886$ and $0.917$, and the
per-setting mean over groups is $0.8995$--$0.9056$ against minima of $0.856$--$0.884$. The
difference between the mean over groups and the minimum, $0.018$--$0.045$, is the figure
Section~\ref{sec:c1} quotes, measured here directly. The group coverages fall on both sides of
$0.90$ in roughly equal numbers, as expected when every group is individually valid.

The shared threshold covers the groups unevenly, and on Waterbirds per-group thresholds narrow the
range across groups to at most $0.031$. On CelebA, the minimum falls in all four settings on an
aligned group holding $47.5\%$ of the calibration set ($g_3$ under ResNet-50 and CLIP, $g_0$ under
DINOv2 and ViT-B/16), while the nominal minority $g_2$ is the \emph{best}-covered group in every
setting. There the coverage ordering runs opposite to the resampled group prior.

\begin{table}[!ht]
\caption{Coverage of each group under both calibration rules (APS, $\rhocal=\rhotest=0.95$), averaged over the kept training methods, their seeds and the ten calibration splits. Groups are ordered $2y+a$, so $g_0$ and $g_3$ are the label-attribute-aligned groups. ``mean'' is the unweighted mean over groups, ``min'' the worst-group coverage reported elsewhere, and ``disparity'' the set-size disparity $\max_g|C|_g-\min_g|C|_g$.}
\label{tab:pergroupcov}
\footnotesize
\setlength{\tabcolsep}{4pt}
\begin{tabular}{@{}llccccccc@{}}
\toprule
 & & \multicolumn{4}{c}{coverage by group} & & & \\
\cmidrule(lr){3-6}
Setting & rule & $g_0$ & $g_1$ & $g_2$ & $g_3$ & mean & min & disparity \\
\midrule
\multicolumn{9}{@{}l}{\emph{Waterbirds}} \\
ResNet-50 & shared & 0.931 & 0.848 & 0.735 & 0.873 & 0.847 & 0.729 & 0.132 \\
 & per-group & 0.901 & 0.911 & 0.914 & 0.893 & 0.905 & 0.870 & 0.278 \\
CLIP & shared & 0.923 & 0.857 & 0.755 & 0.874 & 0.852 & 0.745 & 0.173 \\
 & per-group & 0.903 & 0.917 & 0.898 & 0.886 & 0.901 & 0.856 & 0.244 \\
DINOv2 & shared & 0.908 & 0.896 & 0.866 & 0.885 & 0.889 & 0.850 & 0.080 \\
 & per-group & 0.899 & 0.916 & 0.916 & 0.890 & 0.905 & 0.869 & 0.132 \\
ViT-B/16 & shared & 0.916 & 0.869 & 0.807 & 0.883 & 0.869 & 0.792 & 0.140 \\
 & per-group & 0.900 & 0.915 & 0.914 & 0.893 & 0.906 & 0.869 & 0.206 \\
\multicolumn{9}{@{}l}{\emph{CelebA}} \\
ResNet-50 & shared & 0.925 & 0.944 & 0.953 & 0.877 & 0.925 & 0.865 & 0.339 \\
 & per-group & 0.898 & 0.901 & 0.900 & 0.903 & 0.900 & 0.879 & 0.323 \\
CLIP & shared & 0.907 & 0.920 & 0.933 & 0.894 & 0.913 & 0.866 & 0.239 \\
 & per-group & 0.898 & 0.907 & 0.898 & 0.906 & 0.902 & 0.884 & 0.251 \\
DINOv2 & shared & 0.888 & 0.893 & 0.934 & 0.916 & 0.908 & 0.881 & 0.152 \\
 & per-group & 0.897 & 0.895 & 0.901 & 0.908 & 0.900 & 0.878 & 0.180 \\
ViT-B/16 & shared & 0.884 & 0.905 & 0.924 & 0.913 & 0.906 & 0.881 & 0.163 \\
 & per-group & 0.899 & 0.900 & 0.900 & 0.899 & 0.900 & 0.877 & 0.151 \\
\botrule
\end{tabular}
\end{table}

\input{appendix-tables}

\section{Sub-Target Coverage under Per-Group Thresholds}
\label{app:subtarget}

We ask why worst-group coverage under per-group thresholds, averaged over methods, is
$0.856$--$0.884$ under APS, below the nominal $0.90$ (Section~\ref{sec:c1}). At our calibration sizes, this
is about where an \emph{exactly valid} procedure would put it. The level therefore mainly reflects
the reported statistic and gives no evidence of a validity failure.

Two finite-sample effects combine. The first comes from the statistic we report. Per-group split
conformal with the finite-sample-corrected quantile $\lceil(n_g+1)(1-\alpha)\rceil$ covers at
$\ge1-\alpha$ \emph{in expectation over calibration draws}, but the minimum $\min_g\cov_g$ of
$|\mathcal{G}|$ noisy coverages is biased low. For approximately Gaussian per-group coverages with
standard deviation $\sigma$, $\mathbb{E}[\min_g\cov_g]\approx(1-\alpha)-c_{|\mathcal{G}|}\sigma$
with $c_4\approx1.03$ (Appendix~\ref{app:proofs}).

The second effect is that each per-group threshold rests on that group's \emph{own} calibration
sample, and the samples are unequal. On Waterbirds, the two minority groups hold $2.5\%$ of the
calibration set each, a $19\times$ imbalance in nominal count against the aligned groups. In
distinct images the imbalance is $5$--$12\times$, because the aligned $g_3$ is itself drawn
$4.2\times$ over. On CelebA, counting distinct images reverses the imbalance. The aligned $g_3$ is
blond-and-male, and its nominal $47.5\%$ share is $140$ distinct images drawn about fifty times
each, against $340$--$361$ for the two nominal minorities (Table~\ref{tab:data}). On CelebA, the
least precise threshold therefore belongs to one of the two groups with the largest nominal count.
On both datasets, some threshold rests on far less information than its nominal count suggests.
\citet{ding2023classcond} study this regime and propose clustering classes to pool calibration
data when per-class counts are small.

\begin{figure}[!ht]
\centering
\includegraphics[width=\textwidth]{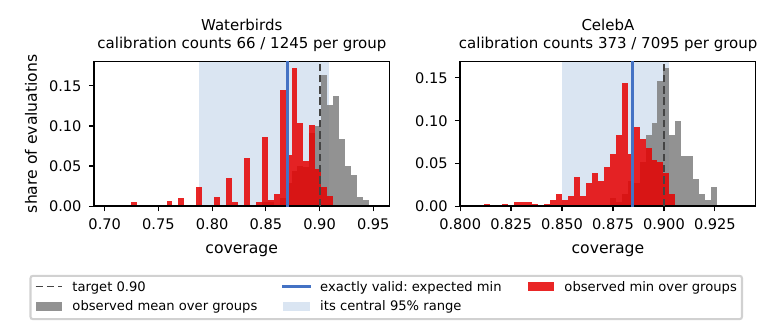}
\caption{The mean over groups (gray) and the minimum over the four groups (red) for every evaluation of the kept runs under per-group thresholds, one panel per dataset (APS, $\rhocal=\rhotest=0.95$). The blue line marks where an exactly valid procedure puts the minimum, simulated at our own per-group counts. Only the locations are compared, not the widths.}
\label{fig:subtarget}
\end{figure}

Because the minimum is not tied to one kind of group (Appendix~\ref{app:proofs}), we use the
simulation below to predict only its level, and Figure~\ref{fig:subtarget} compares our evaluations
with that prediction. Simulated at
our own per-group counts, an exactly valid procedure puts the minimum at $0.8702$ on
Waterbirds and $0.8847$ on CelebA; we observe $0.8660$ and $0.8799$, with $96.5\%$ and
$95.1\%$ of evaluations inside the simulated central range. Both simulations use $3000$ draws and
the nominal per-group counts. Nominal counts understate the noise on the groups where the minimum
tends to fall. The residual of $0.004$--$0.005$ between the simulated and observed minima is
therefore an upper bound on what the simulation leaves unexplained. Table~\ref{tab:subtarget}
repeats the simulation in a fully synthetic setting with four groups, i.i.d.\ continuous scores,
exact exchangeability and per-group conformal thresholds, varying only the realized per-group
calibration counts.

\begin{table}[!ht]
\centering
\caption{Simulated worst-group coverage of an \emph{exactly valid} procedure with per-group thresholds under exchangeability, by calibration-set size ($\rhocal=0.95$; groups ordered $2y+a$; $\alpha=0.10$; $3000$ draws). Per-group counts are the rounded $0.475/0.025$ shares, so a row need not sum to its calibration-set size. The mean over groups approaches $0.90$ from above, because the discrete quantile is conservative when a minority group has only thirteen calibration points. The shortfall falls with the minority count, apart from one reversal between the second and third rows.}
\label{tab:subtarget}
\small
\begin{tabular}{rlccc}
\toprule
cal.\ set & per-group counts & mean over groups & $\mathbb{E}[\min_g\cov_g]$ & shortfall \\
\midrule
$500$    & $238/13/13/238$      & 0.915 & 0.842 & 0.058 \\
$1{,}000$  & $475/25/25/475$      & 0.912 & 0.858 & 0.042 \\
$2{,}000$  & $950/50/50/950$      & 0.901 & 0.856 & 0.044 \\
$5{,}000$  & $2375/125/125/2375$  & 0.902 & 0.874 & 0.026 \\
$10{,}000$ & $4750/250/250/4750$  & 0.900 & 0.880 & 0.020 \\
\botrule
\end{tabular}
\end{table}

The simulated shortfalls, $0.02$--$0.06$, are on the same scale as the observed $0.016$--$0.044$
without quite bracketing them. The observed shortfall also falls with calibration-set size, and it
is larger on Waterbirds than on CelebA, whose calibration sets are much larger.

The shortfall is larger where the per-group thresholds are noisier. Over the three methods whose seeds differ
(Table~\ref{tab:variance}), the split-to-split SD of worst-group coverage under per-group
thresholds is $0.021$--$0.036$ on Waterbirds against $0.011$--$0.019$ on CelebA. The shortfall from
target is likewise $0.029$--$0.043$ against $0.016$--$0.023$. Both follow from sample size. The
minority groups hold a few percent of each calibration set, and CelebA's calibration sets are the
larger ones.

The shortfall comes mainly from reporting a minimum. Measured directly in
Table~\ref{tab:pergroupcov}, the unweighted mean over groups is $0.9024$ against a nominal $0.90$,
and the mean of the per-setting minima is $0.0296$ lower, at $0.8728$. We report $\min_g\cov_g$
because it is the operationally relevant number, and we always give it beside the mean to keep the
two apart.

\section{Pre-Specified Hypotheses}
\label{app:hyps}

We quote the six hypotheses as they were fixed before the experiments they concern
(Section~\ref{sec:exp}), with their original terms: \emph{marginal} or \emph{marginal split} means
one shared threshold, \emph{Mondrian} means per-group thresholds, and \emph{calibration policy}
means the calibration rule. We report each outcome, including the negative ones.

\begin{hypothesis}[Coverage is a calibration property]\label{hyp:c1}
Under Mondrian, worst-group coverage is near target for every training method
including ERM, while under marginal split it falls below target by an amount that
depends on the training method---so the calibration policy governs worst-group
coverage far more than the representation does.
\end{hypothesis}
\begin{hypothesis}[Training buys efficiency]\label{hyp:c2}
Holding Mondrian fixed, worst-group set size differs by training method and tracks
base accuracy.
\end{hypothesis}

\emph{As tested}, set size in Hypothesis~\ref{hyp:c2} is compared with \emph{worst-group}
accuracy, the quantity the training methods are chosen to move, in place of base accuracy
(Table~\ref{tab:c2}). We made this substitution when writing the analysis.

\begin{hypothesis}[Transfer]\label{hyp:h1}
Robust training lowers the accuracy-matched divergence relative to ERM (CI of
\eqref{eq:matched} excludes $0$ on $\ge 2/3$ scores at $\rhocal$).
\end{hypothesis}
\begin{hypothesis}[Ranking inversion]\label{hyp:h2}
The training method best for worst-group accuracy differs from the one best for
the worst group's coverage shortfall, with a CI-separated $\covgap$ gap.
\end{hypothesis}
\begin{hypothesis}[Shift survival]\label{hyp:h3}
Any reduction or coverage level holds across the $\rhotest$ sweep.
\end{hypothesis}
\begin{hypothesis}[No auditability tension]\label{hyp:audit}
The spurious attribute remains recoverable from the representation on which
worst-group coverage is obtained, so coverage and post-hoc group auditing coexist.
\end{hypothesis}

Three of the six hypotheses came back split (Table~\ref{tab:hypverdict}), among them
Hypothesis~\ref{hyp:c1}, whose second clause holds and first does not. The second clause, that
worst-group coverage under one shared threshold falls below target by a training-dependent amount,
is confirmed by the spread across training methods in seven of the eight settings. The exception is
ViT-B/16/CelebA, where the shared threshold already leaves almost nothing to recover. The first
clause, near-target worst-group coverage under per-group thresholds for every method, is
\textbf{not met}. The criterion of Hypothesis~\ref{hyp:c1}, fixed with the hypotheses, requires
\emph{Mondrian within $0.02$ of target for every kept training method, and marginal split
significantly below target for every one of them}, that is, every kept method at $0.88$ or above.
At $\rhotest=0.95$, the average worst-group coverage misses this bar on both datasets, at $0.8799$
on CelebA and $0.866$ on Waterbirds, although six of CelebA's fourteen kept (setting, method) pairs
clear it. Appendix~\ref{app:excl} reports the criterion met in one setting of eight. The
dissociation we report rests on the second clause and on the size of the spreads. It does not
require per-group thresholds to reach target (Appendix~\ref{app:subtarget}).

H\ref{hyp:c2}a in Table~\ref{tab:hypverdict} is marked confirmed because equivalence at a $0.10$
margin is not established in six of eight settings (Section~\ref{sec:c2}). Failing to establish
equivalence is not evidence of a difference. It means only that a set-size spread of $0.10$ or more
cannot be ruled out there.

\begin{table}[!ht]
\caption{Outcomes of the six pre-specified hypotheses. For the three that came back split, we
report the clause that failed beside the one that held, without scoring the hypothesis as a whole.}
\label{tab:hypverdict}
\footnotesize
\setlength{\tabcolsep}{4pt}
\begin{tabular}{@{}l>{\raggedright\arraybackslash}p{3.1cm}>{\raggedright\arraybackslash}p{5.3cm}l@{}}
\toprule
& Predicted & Result & Verdict \\
\midrule
H\ref{hyp:c1}a & Mondrian near target for every method
& Met in one setting of eight on the criterion of Hypothesis~\ref{hyp:c1} (within $0.02$); the level is $0.83$--$0.89$
& \textbf{not met} \\
\addlinespace
H\ref{hyp:c1}b & Marginal split below target, method-dependent
& Spread up to $0.358$ under a shared threshold against at most $0.024$ under per-group in all
eight; the shared spread is the larger in seven, all but ViT-B/16/CelebA & \textbf{confirmed in seven} \\
\addlinespace
H\ref{hyp:c2}a & Set size differs by training method
& $0.022$--$0.259$; equivalence not established at a $0.10$ margin in six of eight
& \textbf{confirmed} \\
\addlinespace
H\ref{hyp:c2}b & and tracks accuracy
& Only one per-setting interval excludes zero, and it rests on three methods; the most accurate
method is the most efficient in four of eight & \textbf{tendency only} \\
\addlinespace
H\ref{hyp:h1} & Robust training lowers the accuracy-matched divergence
& Present in the one setting where model-level supports overlap ($+0.084$, $+0.084$, $+0.150$ across
scores, all intervals excluding zero); undefined in the other seven
& \textbf{met where testable} \\
\addlinespace
H\ref{hyp:h2} & The accuracy-best method differs from the shortfall-best
& Point-estimate inversions in five of eight; one clears its interval, by $0.014$
& \textbf{weakly supported} \\
\addlinespace
H\ref{hyp:h3} & Any reduction or level holds across the $\rhotest$ sweep
& The coverage level holds---under per-group thresholds and APS, worst-group coverage moves by at most $0.024$ over the sweep---while
the matched-divergence reduction becomes undefined & \textbf{split} \\
\addlinespace
H\ref{hyp:audit} & The attribute stays recoverable where coverage is obtained
& Probe $\auroc\in[0.945,0.999]$ in all eight settings while per-group thresholds give $0.83$--$0.89$;
projecting the attribute out moves coverage by at most $0.022$, past the $0.02$ tradeoff criterion only on
CLIP/Waterbirds (Section~\ref{sec:audit}) & \textbf{confirmed, one exception} \\
\botrule
\end{tabular}
\end{table}

\section{Excluded Runs and Sensitivity Analysis}
\label{app:excl}

\paragraph{Accuracy floors.} Table~\ref{tab:floors} gives the floor of each training method. Of the
$33$ excluded runs, AFR on all four CelebA
backbones reaches $0.013$--$0.434$ against its floor of $0.75$, and ERM on CelebA reaches $0.203$
with DINOv2 and $0.263$ with ViT-B/16 features. The other three are individual seeds: GroupDRO-LL
seeds $1$ and $4$ on CelebA/ResNet-50 ($0.676$ and $0.683$) and balanced-subsampling seed $4$ there
($0.730$). We record the reason for every exclusion. The twelve flagged DFR runs on CelebA fall
between DFR's floor of $0.80$ and its soft reference of $0.85$, at $0.808$--$0.847$.
Table~\ref{tab:dagger} gives the two excluded ERM settings in full, to show what the baseline would
have contributed.

\begin{table}[!ht]
\caption{Worst-group accuracy floor of each training method. A run whose worst-group accuracy falls below its floor is excluded from the main analysis.}
\label{tab:floors}
\footnotesize
\setlength{\tabcolsep}{5pt}
\begin{tabular}{@{}lcc@{}}
\toprule
Method & Waterbirds & CelebA \\
\midrule
ERM & 0.30 & 0.30 \\
DFR & 0.45 & 0.80 \\
AFR & 0.40 & 0.75 \\
Balanced subsampling & 0.40 & 0.75 \\
GroupDRO-LL & 0.45 & 0.75 \\
\botrule
\end{tabular}
\end{table}

\paragraph{ERM's worst-group accuracy.} ERM's values below $0.30$, in the two excluded settings,
indicate strong shortcut learning. ERM reaches the robust methods' range of $0.63$--$0.95$ only on
DINOv2/Waterbirds. That is also the Waterbirds setting where the shared threshold came closest to
its target.

\begin{table}[!ht]
\caption{The daggered ERM values of Table~\ref{tab:c1}: worst-group coverage under both calibration rules (APS, $\rhocal=\rhotest=0.95$), the worst-group accuracy that triggers the exclusion, and the floor it misses. Table~\ref{tab:h1} prints dashes for the same settings instead, because an ERM model below its floor cannot serve as the matching reference.}
\label{tab:dagger}
\footnotesize
\setlength{\tabcolsep}{5pt}
\begin{tabular}{@{}lcccc@{}}
\toprule
 & \multicolumn{2}{c}{worst-group coverage} & & \\
\cmidrule(lr){2-3}
Setting & shared & per-group & wg acc & floor \\
\midrule
DINOv2 & 0.813 & 0.885 & 0.203 & 0.30 \\
ViT-B/16 & 0.804 & 0.886 & 0.263 & 0.30 \\
\botrule
\end{tabular}
\end{table}

\paragraph{AFR on CelebA.} AFR's exclusion is a property of the method in this regime, not of our
choice of $\gamma$. AFR reaches $0.686$--$0.890$ worst-group accuracy on Waterbirds but
$0.013$--$0.029$ on the three ViT backbones of CelebA. The cause is a class-prior inversion. At the
default $\gamma=2$, the head predicts the positive class for $\approx\!86\%$ of examples where the
true rate is $\approx\!15\%$, and a \emph{majority} group becomes the worst. At $\gamma=1$, AFR's
worst-group accuracy rises above ERM's on all three but stays far below the floor. Excluding AFR
over an untuned hyperparameter would be unfair, and we therefore computed an \emph{oracle} upper
bound with the best $\gamma$ in $\{0.5,1,2,4\}$ chosen on the evaluation set. No selection rule
over that grid could beat it. The bound is $0.477$ on CLIP, $0.445$ on ViT-B/16, $0.427$ on
ResNet-50 and $0.327$ on DINOv2, all far below the $0.75$ floor. At $\gamma=2$ on ResNet-50, this
separate re-run falls $0.007$ below the $0.434$ of Table~\ref{tab:lit}, a re-run difference of the
size Appendix~\ref{app:hyper} describes.

\paragraph{Restoring the excluded runs.} We ask whether putting the excluded runs back changes the
dissociation, and find that it does not. The excluded runs are individual seeds or whole methods,
and restoring the methods is the informative test. Of the three individual seeds, all on
CelebA/ResNet-50, only GroupDRO-LL seed $1$ falls in the three-seed calibration ablation, where
per-group thresholds are evaluated. The two seed-$4$ runs, of GroupDRO-LL and balanced subsampling,
belong to the five-seed main grid, which uses one shared threshold. Restoring seed $1$ moves its
setting's per-group spread from $0.0034$ to $0.0036$, far below the maximum of $0.024$ on
Waterbirds. The result is therefore arithmetic rather than evidence.

Restoring the whole excluded methods is the larger perturbation. We recomputed every result with
them included. Under per-group thresholds, the spread across training methods rises only to
$0.027$ at worst, against $0.024$ with the methods excluded. Under the shared threshold it grows,
because the degenerate methods are badly under-covered. Only the pre-specified criterion of
Hypothesis~\ref{hyp:c1} changes its outcome. It is met in exactly one of eight settings,
CLIP/CelebA, and that setting flips to failing when the excluded AFR method is included. AFR's
worst-group coverage there is $0.858$ under per-group thresholds, short of the $0.88$ the criterion
asks for, while its $0.695$ under the shared threshold meets the other clause. The criterion is
therefore not robust to the exclusion rule, and we rest the claim on the size of the spreads
rather than on a pass/fail outcome.

\section{Experimental Details}
\label{app:hyper}

\paragraph{Calibration.} Per-group thresholds use one conformal quantile per group, without
pooling. A test point whose group is absent from the calibration set would receive the full label
set, but this never occurs here.

\paragraph{Shift-robust procedure.} Shift-robust calibration uses a worst-case threshold over a
total-variation ball around the calibration distribution, with radius equal to the observed shift.
Unlike the other two rules, it sets its ball radius from test scores at their true labels. It
cannot be deployed, and we report it as an upper bound (Appendix~\ref{app:supp}). Let $\epsilon$ be
the total-variation distance between
histograms, on $50$ shared bins, of the calibration and test scores at the true labels. The
threshold is the $\lceil(n+1)(1-\alpha+\epsilon)\rceil$-th smallest of the $n$ calibration scores,
or $+\infty$ (the full set) when that index exceeds $n$. Raising the level by $\epsilon$ keeps
coverage at least $1-\alpha$ for any test distribution within total-variation distance $\epsilon$ of the
calibration distribution.

\paragraph{Calibration and test construction.} For each of the $10$ calibration-split seeds, a
seeded permutation partitions the evaluation set once into two disjoint halves, the calibration
and test reservoirs. Each reservoir is resampled to its target correlation strength
($\rhocal=0.95$ for calibration, $\rhotest$ for test) by drawing \emph{with replacement} within
each of its groups. Drawing with replacement makes extreme $\rho$ reachable from a finite set of
images. Both draws are truncated to a common size $n=\min(|\mathrm{cal}|,|\mathrm{test}|)$, the
same at every $\rho$, and we record the empirical $\rho$ of each draw. Resampling also balances the
classes. Blond images therefore make up $50\%$ of CelebA's evaluation data, against about $15\%$ in
the raw data. APS and RAPS are randomized and draw their tie-breaking uniforms from separate seeded
streams for calibration and test. RAPS uses $\lambda=0.01$ and $k_{\mathrm{reg}}=1$.
\begin{table}[!ht]
\centering
\caption{Split sizes and group structure. Both datasets are binary with four groups. The ``cal
$=$ test'' and ``per group'' columns give the resampled calibration and test sets at
$\rhocal=\rhotest=0.95$. They are equal in size by the truncation described above, and $\rho$
fixes their group shares at $0.475/0.025/0.025/0.475$ for $2y+a=0,1,2,3$. Because the draw is with
replacement, ``distinct images'' counts the different images behind each group's count.}
\label{tab:data}
\small
\setlength{\tabcolsep}{4pt}
\begin{tabular}{lcccrll}
\toprule
Dataset & $|\mathcal{Y}|$ & $|\mathcal{G}|$ & train & cal $=$ test & per group & distinct images \\
\midrule
Waterbirds & 2 & 4 & 4{,}795 & 2{,}622 & 1245/66/66/1245 & 726/64/59/294 \\
CelebA & 2 & 4 & 162{,}770 & 14{,}936 & 7095/373/373/7095 & 4426/361/340/140 \\
\botrule
\end{tabular}
\end{table}

\paragraph{Per-group counts.} The evaluation sets hold $5{,}245$ images on Waterbirds and $29{,}872$ on CelebA. Over groups
$2y+a=0,1,2,3$, the held-out reweighting split used by DFR and AFR holds $663/706/202/177$ images
on Waterbirds and $4{,}650/3{,}895/1{,}331/81$ on CelebA. The smallest resampled calibration
counts are $66$ on Waterbirds and $373$ on CelebA (Table~\ref{tab:data}), and every per-group
threshold therefore rests on at least $66$ calibration scores.

\paragraph{Nominal and distinct counts.} With replacement, a group's resampled count can exceed
its number of distinct images, and Table~\ref{tab:data} gives both. The minority groups are
essentially subsampled, at $1.0$--$1.1$ draws per distinct image. The aligned groups are inflated,
mildly on Waterbirds ($1{,}245$ draws of $g_3$ from $294$ images) and severely on CelebA ($7{,}095$
draws of the blond-and-male $g_3$ from $140$ images, Appendix~\ref{app:subtarget}). We evaluate
both calibration rules on the same resampled data, and the comparison between them is therefore
unaffected. The split-to-split variation in Table~\ref{tab:variance} reflects the repeated draws.

\paragraph{Heads, backbone and fine-tuning.} Table~\ref{tab:config} gives the training
configuration. Every head standardizes features with the statistics of the split it is fitted on,
and all heads except GroupDRO-LL are multinomial logistic models. Appendix~\ref{app:excl} examines
AFR's sensitivity to $\gamma$. The frozen ResNet-50 starts from the torchvision
\texttt{IMAGENET1K\_V2} checkpoint and is trained in-domain before its features are fixed.
Fine-tuning (Section~\ref{sec:repr}) also starts from the \texttt{IMAGENET1K\_V2} checkpoint, uses
batch size $128$ and mixed precision, and keeps the final epoch without checkpoint selection.

\begin{table}[!ht]
\caption{Training configuration of the heads, the frozen ResNet-50 backbone and the three fine-tuning objectives of Section~\ref{sec:repr}.}
\label{tab:config}
\footnotesize
\setlength{\tabcolsep}{4pt}
\begin{tabular}{@{}>{\raggedright\arraybackslash}p{2.6cm}>{\raggedright\arraybackslash}p{3.3cm}>{\raggedright\arraybackslash}p{6.4cm}@{}}
\toprule
 & Fitted on & Settings \\
\midrule
\multicolumn{3}{@{}l}{\emph{Heads}} \\
ERM & training split & L-BFGS, $C=1$, at most $5000$ iterations \\
DFR & reweighting split & L-BFGS, $C=1$, at most $5000$ iterations; averages $10$ group-balanced subsamples \\
AFR & reweighting split & L-BFGS, $C=1$, at most $5000$ iterations; $\gamma=2$ \\
Balanced subsampling & training split & L-BFGS, $C=1$, at most $5000$ iterations; draws one group-balanced subsample \\
GroupDRO-LL & training split & $2000$ full-batch steps at learning rate $0.05$, group-weight step $\eta_q=0.05$, $\ell_2$ penalty $10^{-3}$ \\
\addlinespace
\multicolumn{3}{@{}l}{\emph{Frozen backbone}} \\
ResNet-50 & full training split (Waterbirds); random subsample of $30{,}000$ training images (CelebA) & ERM for $10$ epochs, Adam at $10^{-3}$, batch size $128$ \\
\addlinespace
\multicolumn{3}{@{}l}{\emph{Fine-tuning}} \\
ERM & \multicolumn{2}{@{}l}{Adam at $10^{-3}$; $10$ epochs on Waterbirds, $6$ on CelebA} \\
Reweighting & \multicolumn{2}{@{}l}{samples groups at inverse frequency; Adam at $10^{-3}$; $10$ epochs on Waterbirds, $6$ on CelebA} \\
GroupDRO & \multicolumn{2}{@{}l}{SGD at $10^{-3}$, momentum $0.9$, weight decay $10^{-2}$, $\eta_q=0.05$; $20$ epochs on Waterbirds, $6$ on CelebA} \\
\botrule
\end{tabular}
\end{table}

\paragraph{Comparison with published values.} Two protocol differences separate our ResNet-50
methods from the published values of Table~\ref{tab:lit} (Section~\ref{sec:exp}). The published
GroupDRO numbers come from end-to-end training, and the published DFR and AFR numbers retrain a last
layer on a stronger base model. DFR comes closest ($0.851$ against $0.883$ on CelebA). AFR, which
reweights the base model's own errors, is furthest on CelebA, where it degenerates
(Appendix~\ref{app:excl}). On Waterbirds, last-layer GroupDRO is furthest. The gaps do not affect
our main conclusions. These rest on comparisons within a setting on the same evaluation data, and
the fine-tuning study (Section~\ref{sec:repr}) and the projection test in Section~\ref{sec:h3}
change the features by design. A weaker backbone lowers all methods in a setting together. The four
backbones span $0.20$--$0.95$ in worst-group accuracy and show the same pattern.

\paragraph{Matching level.} Pooling each method's $(\mathrm{acc},D)$ points across seeds
\emph{and} calibration splits would make the accuracy axis mostly evaluation-draw noise. ERM's head
is the same for every seed (Section~\ref{sec:exp}), and its pooled accuracies differ only in which
evaluation images were drawn. A bootstrap over the pooled points would also treat draws from one
trained model as independent. We therefore match at model level (Section~\ref{sec:h1}). Six pairs
whose pooled accuracies overlap ERM's at split level, among them GroupDRO-LL on CLIP/Waterbirds and
balanced subsampling on ResNet-50/Waterbirds, do not overlap at model level. Accuracy is
overall (base) top-1 accuracy on the evaluation distribution. Each robust method's $D(a^\star)$
comes from a least-squares line through its per-seed points. ERM's is its mean $D$, since its
accuracy is a single point. On the two CelebA settings with an ERM reference, the robust methods
raise base accuracy to $0.81$--$0.91$ against ERM's $0.68$--$0.70$ (Table~\ref{tab:h1}). On the
three Waterbirds settings without a matched pair, their mean accuracies run from $0.149$ below ERM's
to $0.001$ above, but no method's support contains ERM's accuracy.

\begin{table}[!ht]
\centering
\caption{Divergence and base accuracy for ERM and the robust methods, from the five-seed main grid (APS, $\rhocal=0.95$). The $D$ and accuracy columns give ranges of method means, so they need not agree with the last column, which is assessed per training seed at model level (Section~\ref{sec:h1}). It names the methods whose accuracy support overlaps ERM's; \texttt{unm.}\ means none does, so $\Delta(a^\star)$ is undefined. $^{\dagger}$ERM excluded (Appendix~\ref{app:excl}).}
\label{tab:h1}
\footnotesize
\setlength{\tabcolsep}{4pt}
\begin{tabular}{lccccc}
\toprule
Backbone & $D_{\mathrm{ERM}}$ & acc$_{\mathrm{ERM}}$ & $D$ (rob.) & acc (rob.) & matched \\
\midrule
\multicolumn{6}{l}{\emph{Waterbirds}} \\
ResNet-50 & 0.190 & 0.940 & 0.038--0.110 & 0.800--0.939 & GroupDRO-LL \\
CLIP & 0.216 & 0.932 & 0.036--0.118 & 0.783--0.928 & \texttt{unm.} \\
DINOv2 & 0.053 & 0.983 & 0.031--0.040 & 0.958--0.982 & \texttt{unm.} \\
ViT-B/16 & 0.141 & 0.944 & 0.033--0.089 & 0.845--0.945 & \texttt{unm.} \\
\midrule
\multicolumn{6}{l}{\emph{CelebA}} \\
ResNet-50 & 0.190 & 0.699 & 0.013--0.031 & 0.810--0.859 & \texttt{unm.} \\
CLIP & 0.197 & 0.677 & 0.023--0.030 & 0.900--0.913 & \texttt{unm.} \\
DINOv2$^{\dagger}$ & --- & --- & 0.014--0.032 & 0.830--0.876 & \texttt{no ref.} \\
ViT-B/16$^{\dagger}$ & --- & --- & 0.022--0.033 & 0.833--0.893 & \texttt{no ref.} \\
\botrule
\end{tabular}
\end{table}

\paragraph{Recoverability.} The projection test of Section~\ref{sec:audit} removes directions one
at a time from both training and evaluation features. Each is the normalized weight of a logistic
probe refitted on the training features left after the previous removals.

\paragraph{Seeds.} Training seeds are $\{0,1,2,3,4\}$ for the main grid and for the fine-tuning
study on Waterbirds, and $\{0,1,2\}$ for the fine-tuning study on CelebA, the calibration ablation
and the predicted-group study. Calibration-split seeds are $0$--$9$ throughout, and bootstrap
seeds are fixed at $0$. ERM and AFR produce the same head for every seed, with base accuracy
identical to five decimals in every setting (Section~\ref{sec:exp}).

\paragraph{Compute.} Features are extracted once per backbone. The main grid, the calibration
ablation and the predicted-group study then reuse precomputed head posteriors, and each takes a
few hours on a single machine, with no GPU retraining.

\paragraph{Reproducibility across machines.} We re-ran the coverage evaluation behind
Table~\ref{tab:pergroupcov} on a second machine, with the same extracted features and released
splits. DFR and GroupDRO-LL reproduce exactly, while ERM, AFR and balanced subsampling differ in
$305$ of the $2{,}400$ evaluations ($192$ of them AFR's), by at most $0.013$ in worst-group coverage.
Their L-BFGS solves run at full precision and are deterministic on either machine, but under a
different linear-algebra build a solve can stop at a solution that differs in the last bit and move
a few test points across a threshold, twice over for AFR, whose first head's probabilities weight
its second. All reported values, set sizes included, are the released ones, their per-setting means
agree with the re-run to within $0.0004$, and a reader re-running these three methods should expect
the third decimal to change, occasionally the second.

\paragraph{Archival code.} The repository listed under ``Data and code availability'' holds the
code that produces each table and figure, one script per experimental stage, and the exact commands
that reproduce every released record from the extracted features. Its DOI resolves to the latest
release.

\bibliographystyle{plainnat}
\bibliography{sn-bibliography}

\end{document}

%% file: appendix-tables.tex
\begin{table}[!ht]
\centering
\caption{The dissociation under all three scores ($\rhocal=\rhotest=0.95$). Each entry is the spread in worst-group coverage across the kept training methods, under one shared threshold or under per-group thresholds.}
\label{tab:scorespread}
\footnotesize
\setlength{\tabcolsep}{4pt}
\begin{tabular}{@{}lcccccc@{}}
\toprule
 & \multicolumn{2}{c}{APS} & \multicolumn{2}{c}{RAPS} & \multicolumn{2}{c}{THR} \\
\cmidrule(lr){2-3}\cmidrule(lr){4-5}\cmidrule(lr){6-7}
Setting & shared & per-gr. & shared & per-gr. & shared & per-gr. \\
\midrule
\multicolumn{7}{@{}l}{\emph{Waterbirds}} \\
ResNet-50 & 0.290 & 0.008 & 0.297 & 0.008 & 0.464 & 0.024 \\
CLIP & 0.358 & 0.024 & 0.369 & 0.029 & 0.478 & 0.018 \\
DINOv2 & 0.049 & 0.024 & 0.051 & 0.023 & 0.340 & 0.007 \\
ViT-B/16 & 0.217 & 0.007 & 0.221 & 0.010 & 0.409 & 0.011 \\
\multicolumn{7}{@{}l}{\emph{CelebA}} \\
ResNet-50 & 0.071 & 0.003 & 0.075 & 0.004 & 0.070 & 0.014 \\
CLIP & 0.059 & 0.003 & 0.066 & 0.003 & 0.087 & 0.008 \\
DINOv2 & 0.013 & 0.003 & 0.014 & 0.004 & 0.013 & 0.008 \\
ViT-B/16 & 0.003 & 0.009 & 0.002 & 0.008 & 0.004 & 0.001 \\
\botrule
\end{tabular}
\end{table}

\begin{table}[!ht]
\centering
\caption{Cross-group score divergences under one shared threshold ($\rhocal=\rhotest=0.95$), pooled over all kept runs, including the seven flagged DFR runs on CelebA that Figure~\ref{fig:mech} leaves out. $D_{\mathrm{KS}}$ is the quantity bounded in Eq.~\eqref{eq:ksbound}. $\Wone$, given for APS alone, is a descriptive summary and enters no bound.}
\label{tab:div}
\footnotesize
\setlength{\tabcolsep}{3.5pt}
\begin{tabular}{lcccc}
\toprule
Setting & $D_{\mathrm{KS}}$ APS & $D_{\mathrm{KS}}$ RAPS & $D_{\mathrm{KS}}$ THR & $\Wone$ APS \\
\midrule
\multicolumn{5}{l}{\emph{Waterbirds}} \\
ResNet-50 & 0.214 & 0.217 & 0.361 & 0.103 \\
CLIP & 0.201 & 0.204 & 0.321 & 0.098 \\
DINOv2 & 0.098 & 0.098 & 0.299 & 0.040 \\
ViT-B/16 & 0.157 & 0.156 & 0.291 & 0.071 \\
\multicolumn{5}{l}{\emph{CelebA}} \\
ResNet-50 & 0.130 & 0.133 & 0.533 & 0.070 \\
CLIP & 0.125 & 0.126 & 0.360 & 0.069 \\
DINOv2 & 0.054 & 0.055 & 0.245 & 0.026 \\
ViT-B/16 & 0.050 & 0.050 & 0.277 & 0.026 \\
\botrule
\end{tabular}
\end{table}

\begin{table}[!ht]
\centering
\caption{The fine-tuning study of Section~\ref{sec:repr} for APS and THR ($\rhocal=\rhotest=0.95$), with the head held fixed at plain ERM: means over five fine-tuning seeds on Waterbirds and three on CelebA, with the across-seed standard deviation in brackets. Worst-group accuracy does not depend on the score. RAPS is within $0.005$ of APS throughout.}
\label{tab:reprfull}
\footnotesize
\setlength{\tabcolsep}{3.5pt}
\begin{tabular}{llccc}
\toprule
Objective & Score & wg acc & wg cov shared & wg cov per-group \\
\midrule
\multicolumn{5}{l}{\emph{Waterbirds}} \\
ERM & APS & 0.516 & 0.568\,[0.048] & 0.865\,[0.008] \\
 & THR & 0.516 & 0.410\,[0.055] & 0.865\,[0.002] \\
GroupDRO & APS & 0.738 & 0.733\,[0.012] & 0.861\,[0.007] \\
 & THR & 0.738 & 0.480\,[0.040] & 0.862\,[0.006] \\
Reweighting & APS & 0.466 & 0.485\,[0.033] & 0.864\,[0.006] \\
 & THR & 0.466 & 0.371\,[0.026] & 0.862\,[0.008] \\
\multicolumn{5}{l}{\emph{CelebA}} \\
ERM & APS & 0.396 & 0.805\,[0.007] & 0.882\,[0.005] \\
 & THR & 0.396 & 0.806\,[0.009] & 0.880\,[0.000] \\
GroupDRO & APS & 0.470 & 0.824\,[0.002] & 0.881\,[0.002] \\
 & THR & 0.470 & 0.815\,[0.005] & 0.879\,[0.002] \\
Reweighting & APS & 0.364 & 0.806\,[0.010] & 0.881\,[0.005] \\
 & THR & 0.364 & 0.810\,[0.014] & 0.882\,[0.002] \\
\botrule
\end{tabular}
\end{table}

\begin{table}[!ht]
\centering
\caption{Predicted-group calibration: worst-group coverage with (a) true groups, (b) predicted groups at test only and (c) predicted groups at both calibration and test, as in Table~\ref{tab:predgroup} ($\rhocal=\rhotest=0.95$), averaged over the kept training methods, for APS and THR. Coverage is always scored against the true groups. RAPS is within $0.001$ of APS in every entry.}
\label{tab:pgfull}
\footnotesize
\setlength{\tabcolsep}{3.5pt}
\begin{tabular}{llcccc}
\toprule
Setting & Score & (a) true & (b) pred.\ test & (c) pred.\ both & probe $\auroc$ \\
\midrule
\multicolumn{6}{l}{\emph{Waterbirds}} \\
ResNet-50 & APS & 0.870 & 0.866 & 0.870 & 0.945 \\
 & THR & 0.862 & 0.852 & 0.867 & 0.945 \\
CLIP & APS & 0.856 & 0.855 & 0.845 & 0.976 \\
 & THR & 0.860 & 0.858 & 0.842 & 0.976 \\
DINOv2 & APS & 0.869 & 0.868 & 0.858 & 0.961 \\
 & THR & 0.878 & 0.867 & 0.865 & 0.961 \\
ViT-B/16 & APS & 0.869 & 0.864 & 0.873 & 0.964 \\
 & THR & 0.868 & 0.860 & 0.865 & 0.964 \\
\multicolumn{6}{l}{\emph{CelebA}} \\
ResNet-50 & APS & 0.879 & 0.861 & 0.877 & 0.962 \\
 & THR & 0.876 & 0.838 & 0.870 & 0.962 \\
CLIP & APS & 0.884 & 0.884 & 0.873 & 0.999 \\
 & THR & 0.884 & 0.883 & 0.872 & 0.999 \\
DINOv2 & APS & 0.878 & 0.878 & 0.869 & 0.999 \\
 & THR & 0.879 & 0.879 & 0.879 & 0.999 \\
ViT-B/16 & APS & 0.877 & 0.878 & 0.877 & 0.993 \\
 & THR & 0.869 & 0.866 & 0.866 & 0.993 \\
\botrule
\end{tabular}
\end{table}

\begin{table}[!ht]
\centering
\caption{Standard deviation of worst-group coverage across training seeds and across calibration splits (APS, $\rhocal=\rhotest=0.95$), averaged over the kept training methods. The first two columns use per-group thresholds and the last uses one shared threshold. ERM and AFR are omitted because their solver ignores the seed.}
\label{tab:variance}
\footnotesize
\setlength{\tabcolsep}{3.5pt}
\begin{tabular}{lccc}
\toprule
Setting & SD across seeds & SD across splits & SD across splits \\
 & (per-group) & (per-group) & (shared) \\
\midrule
\multicolumn{4}{l}{\emph{Waterbirds}} \\
ResNet-50 & 0.0027 & 0.0259 & 0.0434 \\
CLIP & 0.0061 & 0.0360 & 0.0305 \\
DINOv2 & 0.0022 & 0.0237 & 0.0205 \\
ViT-B/16 & 0.0019 & 0.0211 & 0.0322 \\
\multicolumn{4}{l}{\emph{CelebA}} \\
ResNet-50 & 0.0011 & 0.0187 & 0.0214 \\
CLIP & 0.0020 & 0.0109 & 0.0074 \\
DINOv2 & 0.0025 & 0.0142 & 0.0119 \\
ViT-B/16 & 0.0019 & 0.0139 & 0.0085 \\
\botrule
\end{tabular}
\end{table}

%% file: sn-bibliography.bib
@inproceedings{
angelopoulos2020,
title={Uncertainty Sets for Image Classifiers using Conformal Prediction},
author={Anastasios Nikolas Angelopoulos and Stephen Bates and Michael Jordan and Jitendra Malik},
booktitle={International Conference on Learning Representations},
year={2021}
}

@inproceedings{
cresswell2024,
title={Conformal Prediction Sets Can Cause Disparate Impact},
author={Jesse C. Cresswell and Bhargava Kumar and Yi Sui and Mouloud Belbahri},
booktitle={The Thirteenth International Conference on Learning Representations},
year={2025}
}

@inproceedings{deng2025kandinsky,
title={Kandinsky Conformal Prediction: Beyond Class- and Covariate-Conditional Coverage},
author={Konstantina Bairaktari and Jiayun Wu and Steven Wu},
booktitle={Forty-second International Conference on Machine Learning},
year={2025}
}

@inproceedings{afcp2024,
title={Conformal Classification with Equalized Coverage for Adaptively Selected Groups},
author={Yanfei Zhou and Matteo Sesia},
booktitle={The Thirty-eighth Annual Conference on Neural Information Processing Systems},
year={2024}
}

@article{gibbs2023,
      title={Conformal Prediction With Conditional Guarantees}, 
      author={Isaac Gibbs and John J. Cherian and Emmanuel J. Candès},
      journal={Journal of the Royal Statistical Society Series B: Statistical Methodology},
      volume={87},
      number={4},
      pages={1100--1126},
      year={2025},
      doi={10.1093/jrsssb/qkaf008}
}

@inproceedings{jones2021,
title={Selective Classification Can Magnify Disparities Across Groups},
author={Erik Jones and Shiori Sagawa and Pang Wei Koh and Ananya Kumar and Percy Liang},
booktitle={International Conference on Learning Representations},
year={2021}
}

@inproceedings{jung2023,
title={Batch Multivalid Conformal Prediction},
author={Christopher Jung and Georgy Noarov and Ramya Ramalingam and Aaron Roth},
booktitle={International Conference on Learning Representations},
year={2023}
}

@inproceedings{kirichenko2023,
title={Last Layer Re-Training is Sufficient for Robustness to Spurious Correlations},
author={Polina Kirichenko and Pavel Izmailov and Andrew Gordon Wilson},
booktitle={The Eleventh International Conference on Learning Representations },
year={2023}
}

@inproceedings{liu2021jtt,
  title = 	 {Just Train Twice: Improving Group Robustness without Training Group Information},
  author =       {Liu, Evan Z and Haghgoo, Behzad and Chen, Annie S and Raghunathan, Aditi and Koh, Pang Wei and Sagawa, Shiori and Liang, Percy and Finn, Chelsea},
  booktitle = 	 {Proceedings of the 38th International Conference on Machine Learning},
  pages = 	 {6781--6792},
  year = 	 {2021},
  editor = 	 {Meila, Marina and Zhang, Tong},
  volume = 	 {139},
  series = 	 {Proceedings of Machine Learning Research},
  month = 	 {18--24 Jul},
  publisher =    {PMLR}
}

@inproceedings{mdcp2026,
      title={Multi-Distribution Robust Conformal Prediction}, 
      author={Yuqi Yang and Ying Jin},
      booktitle={International Conference on Machine Learning (ICML)},
      year={2026}
}

@misc{burden2026,
      title={On the Burden of Achieving Fairness in Conformal Prediction}, 
      author={Ziang Gao and Pengqi Liu and Archer Yi Yang and Mouloud Belbahri and Jesse C. Cresswell and Masoud Asgharian},
      year={2026},
      eprint={2605.14260},
      archivePrefix={arXiv},
      primaryClass={stat.ML}
}

@inproceedings{qiu2023,
  title = 	 {Simple and Fast Group Robustness by Automatic Feature Reweighting},
  author =       {Qiu, Shikai and Potapczynski, Andres and Izmailov, Pavel and Wilson, Andrew Gordon},
  booktitle = 	 {Proceedings of the 40th International Conference on Machine Learning},
  pages = 	 {28448--28467},
  year = 	 {2023},
  editor = 	 {Krause, Andreas and Brunskill, Emma and Cho, Kyunghyun and Engelhardt, Barbara and Sabato, Sivan and Scarlett, Jonathan},
  volume = 	 {202},
  series = 	 {Proceedings of Machine Learning Research},
  month = 	 {23--29 Jul},
  publisher =    {PMLR}
}

@inproceedings{romano2020,
author = {Romano, Yaniv and Sesia, Matteo and Cand\`{e}s, Emmanuel J.},
title = {Classification with valid and adaptive coverage},
year = {2020},
isbn = {9781713829546},
publisher = {Curran Associates Inc.},
address = {Red Hook, NY, USA},
booktitle = {Proceedings of the 34th International Conference on Neural Information Processing Systems},
articleno = {302},
numpages = {11},
location = {Vancouver, BC, Canada},
series = {NIPS '20}
}

@inproceedings{sagawa2020,
title={Distributionally Robust Neural Networks},
author={Shiori Sagawa and Pang Wei Koh and Tatsunori B. Hashimoto and Percy Liang},
booktitle={International Conference on Learning Representations},
year={2020}
}

@inproceedings{tibshirani2019,
author = {Tibshirani, Ryan J. and Barber, Rina Foygel and Cand\`{e}s, Emmanuel J. and Ramdas, Aaditya},
title = {Conformal prediction under covariate shift},
booktitle = {Advances in Neural Information Processing Systems},
volume = {32},
year = {2019}
}

@book{vovk2005,
  author = {Vovk, Vladimir and Gammerman, Alexander and Shafer, Glenn},
  title = {Algorithmic Learning in a Random World},
  publisher = {Springer},
  year = {2005}
}

@inproceedings{yang2023change,
author = {Yang, Yuzhe and Zhang, Haoran and Katabi, Dina and Ghassemi, Marzyeh},
title = {Change is hard: a closer look at subpopulation shift},
year = {2023},
publisher = {JMLR.org},
booktitle = {Proceedings of the 40th International Conference on Machine Learning},
articleno = {1652},
numpages = {39},
location = {Honolulu, Hawaii, USA},
series = {ICML'23}
}

@inproceedings{zhang2022cnc,
  author = {Zhang, Michael and Sohoni, Nimit S. and Zhang, Hongyang R. and Finn, Chelsea and R\'e, Christopher},
  title = {Correct-n-contrast: a contrastive approach for improving robustness to spurious correlations},
  booktitle = {International Conference on Machine Learning (ICML)},
  year = {2022}
}

@article{barber2021limits,
  title={The limits of distribution-free conditional predictive inference},
  author={Rina Foygel Barber and Emmanuel J. Cand{\`e}s and Aaditya Ramdas and Ryan J. Tibshirani},
  journal={Information and Inference: A Journal of the IMA},
  volume={10},
  number={2},
  pages={455--482},
  year={2021}
}

@article{einbinder2022,
  author  = {Bat-Sheva Einbinder and Shai Feldman and Stephen Bates and Anastasios N. Angelopoulos and Asaf Gendler and Yaniv Romano},
  title   = {Label Noise Robustness of Conformal Prediction},
  journal = {Journal of Machine Learning Research},
  year    = {2024},
  volume  = {25},
  number  = {328},
  pages   = {1--66},
}

@article{sesia2023noisy,
      title={Adaptive conformal classification with noisy labels}, 
      author={Matteo Sesia and Y. X. Rachel Wang and Xin Tong},
      journal={Journal of the Royal Statistical Society Series B: Statistical Methodology},
      volume={87},
      number={3},
      pages={796--815},
      year={2025},
      doi={10.1093/jrsssb/qkae114}
}

@article{elmehdi2024unfreezing,
      author={El Moudden, T. E. and Amnai, M. and Choukri, A. and Fakhri, Y. and
              Noreddine, G.},
      title={New unfreezing strategy of transfer learning in satellite imagery for mapping
             the diversity of slum areas: A case study in Kenitra city---Morocco},
      journal={Scientific African},
      volume={23},
      pages={e02135},
      year={2024},
      doi={10.1016/j.sciaf.2024.e02135}
}

@inproceedings{kumar2022finetuning,
  title={Fine-Tuning can Distort Pretrained Features and Underperform Out-of-Distribution},
  author={Ananya Kumar and Aditi Raghunathan and Robbie Matthew Jones and Tengyu Ma and Percy Liang},
  booktitle={International Conference on Learning Representations},
  year={2022}
}

@article{romano2020equalized,
  title={With Malice Toward None: Assessing Uncertainty via Equalized Coverage},
  author={Romano, Yaniv and Barber, Rina Foygel and Sabatti, Chiara and Cand\`{e}s, Emmanuel J.},
  journal={Harvard Data Science Review},
  volume={2},
  number={2},
  year={2020},
  doi={10.1162/99608f92.03f00592}
}

@inproceedings{idrissi2022balancing,
  title={Simple data balancing achieves competitive worst-group-accuracy},
  author={Youbi Idrissi, Badr and Arjovsky, Martin and Pezeshki, Mohammad and Lopez-Paz, David},
  booktitle={Proceedings of the First Conference on Causal Learning and Reasoning},
  series={Proceedings of Machine Learning Research},
  volume={177},
  pages={336--351},
  year={2022},
  publisher={PMLR}
}

@inproceedings{ding2023classcond,
  title={Class-Conditional Conformal Prediction with Many Classes},
  author={Ding, Tiffany and Angelopoulos, Anastasios N. and Bates, Stephen and
          Jordan, Michael I. and Tibshirani, Ryan J.},
  booktitle={Advances in Neural Information Processing Systems},
  volume={36},
  year={2023}
}

@inproceedings{vovk2003mondrian,
  title={Mondrian confidence machine},
  author={Vovk, Vladimir and Lindsay, David and Nouretdinov, Ilia and Gammerman, Alexander},
  booktitle={On-line Compression Modelling Project, Working Paper 4},
  year={2003}
}

@inproceedings{guo2017calibration,
  title={On Calibration of Modern Neural Networks},
  author={Guo, Chuan and Pleiss, Geoff and Sun, Yu and Weinberger, Kilian Q.},
  booktitle={Proceedings of the 34th International Conference on Machine Learning},
  series={Proceedings of Machine Learning Research},
  volume={70},
  pages={1321--1330},
  year={2017},
  publisher={PMLR}
}

@article{lei2014distribution,
  title={Distribution-free prediction bands for non-parametric regression},
  author={Lei, Jing and Wasserman, Larry},
  journal={Journal of the Royal Statistical Society: Series B},
  volume={76},
  number={1},
  pages={71--96},
  year={2014}
}

@InProceedings{he2016resnet,
author = {He, Kaiming and Zhang, Xiangyu and Ren, Shaoqing and Sun, Jian},
title = {Deep Residual Learning for Image Recognition},
booktitle = {Proceedings of the IEEE Conference on Computer Vision and Pattern Recognition (CVPR)},
month = {June},
year = {2016}
}

@InProceedings{liu2015celeba,
author = {Liu, Ziwei and Luo, Ping and Wang, Xiaogang and Tang, Xiaoou},
title = {Deep Learning Face Attributes in the Wild},
booktitle = {Proceedings of the IEEE International Conference on Computer Vision (ICCV)},
month = {December},
year = {2015}
}

@InProceedings{radford2021clip,
  title = {Learning Transferable Visual Models From Natural Language Supervision},
  author = {Radford, Alec and Kim, Jong Wook and Hallacy, Chris and Ramesh, Aditya and Goh, Gabriel and Agarwal, Sandhini and Sastry, Girish and Askell, Amanda and Mishkin, Pamela and Clark, Jack and Krueger, Gretchen and Sutskever, Ilya},
  booktitle = {Proceedings of the 38th International Conference on Machine Learning},
  pages = {8748--8763},
  year = {2021},
  editor = {Meila, Marina and Zhang, Tong},
  volume = {139},
  series = {Proceedings of Machine Learning Research},
  month = {18--24 Jul},
  publisher = {PMLR},
  url = {https://proceedings.mlr.press/v139/radford21a.html}
}

@article{oquab2024dinov2,
title={{DINO}v2: Learning Robust Visual Features without Supervision},
author={Maxime Oquab and Timoth{\'e}e Darcet and Th{\'e}o Moutakanni and Huy V. Vo and Marc Szafraniec and Vasil Khalidov and Pierre Fernandez and Daniel Haziza and Francisco Massa and Alaaeldin El-Nouby and Mido Assran and Nicolas Ballas and Wojciech Galuba and Russell Howes and Po-Yao Huang and Shang-Wen Li and Ishan Misra and Michael Rabbat and Vasu Sharma and Gabriel Synnaeve and Hu Xu and Herve Jegou and Julien Mairal and Patrick Labatut and Armand Joulin and Piotr Bojanowski},
journal={Transactions on Machine Learning Research},
issn={2835-8856},
year={2024},
url={https://openreview.net/forum?id=a68SUt6zFt}
}

@inproceedings{dosovitskiy2021vit,
title={An Image is Worth 16x16 Words: Transformers for Image Recognition at Scale},
author={Alexey Dosovitskiy and Lucas Beyer and Alexander Kolesnikov and Dirk Weissenborn and Xiaohua Zhai and Thomas Unterthiner and Mostafa Dehghani and Matthias Minderer and Georg Heigold and Sylvain Gelly and Jakob Uszkoreit and Neil Houlsby},
booktitle={International Conference on Learning Representations},
year={2021}
}
